\documentclass[10pt,twocolumn,letterpaper]{article}

\usepackage{iccv}
\usepackage{times}
\usepackage{epsfig}
\usepackage{graphicx}
\usepackage{amsmath}
\usepackage{amssymb}
\usepackage{gensymb}
\usepackage{epstopdf}
\usepackage{makecell}
\usepackage{multirow}
\usepackage{setspace}
\usepackage{graphics}
\usepackage{subfigure}
\usepackage{appendix}

\usepackage[accsupp]{axessibility}

\usepackage[pagebackref=true,breaklinks=true,letterpaper=true,colorlinks,bookmarks=false]{hyperref}

\iccvfinalcopy 

\def\iccvPaperID{7732} 

\ificcvfinal\fi

\begin{document}

\title{ADNet: Leveraging Error-Bias Towards Normal Direction in Face Alignment}

\author{Yangyu Huang,~Hao Yang\thanks{Corresponding author}~,~Chong Li,~Jongyoo Kim,~Fangyun Wei\\
Microsoft Research Asia\\
{\tt\small \{yanghuan,haya,chol,jongk,fawe\}@microsoft.com}
}

\maketitle
\ificcvfinal\thispagestyle{empty}\fi

\begin{abstract}
  The recent progress of CNN has dramatically improved face alignment performance. 
  However, few works have paid attention to the error-bias with respect to error distribution of facial landmarks. In this paper, we investigate the error-bias issue in face alignment, where the distributions of landmark errors tend to spread along the tangent line to landmark curves.
  This error-bias is not trivial since it is closely connected to the ambiguous landmark labeling task. Inspired by this observation, we seek a way to leverage the error-bias property for better convergence of CNN model. 
  To this end, we propose anisotropic direction loss (ADL) and anisotropic attention module (AAM) for coordinate and heatmap regression, respectively.
  ADL imposes strong binding force in normal direction for each landmark point on facial boundaries.
  On the other hand, AAM is an attention module which can get anisotropic attention mask focusing on the region of point and its local edge connected by adjacent points, it has a stronger response in tangent than in normal, which means relaxed constraints in the tangent. These two methods work in a complementary manner to learn both facial structures and texture details.
  Finally, we integrate them into an optimized end-to-end training pipeline named ADNet. Our ADNet achieves state-of-the-art results on 300W, WFLW and COFW datasets, which demonstrates the effectiveness and robustness.
\end{abstract}

\section{Introduction}
\label{section:introduction}
Face alignment, applied to facial landmark detection, has experienced tremendous improvement by means of Convolutional Neural Networks, and provides a continuous impetus for improvements in many computer vision techniques for face such as face recognition \cite{masi2018deep}, face synthesis \cite{bao2018towards,gu2019mask} and face 3D reconstruction \cite{jiang2005efficient}.

Error-bias can be treated as a special kind of AI-bias that could be resulted from the prejudiced assumptions made in the process of algorithm development or prejudices in the training data thus is an anomaly in the output of machine learning algorithms. Common AI-bias, such as race-bias and gender-bias, always causes negative ethical effects, especially in the case of gender shades \cite{buolamwini2018gender}. Different from them, the error-bias considered in this paper is more like a location uncertainty, as mentioned by \cite{kumar2020luvli}.

\begin{figure}[t]
\centering
\subfigure[Human]
{\includegraphics[width=0.93in]{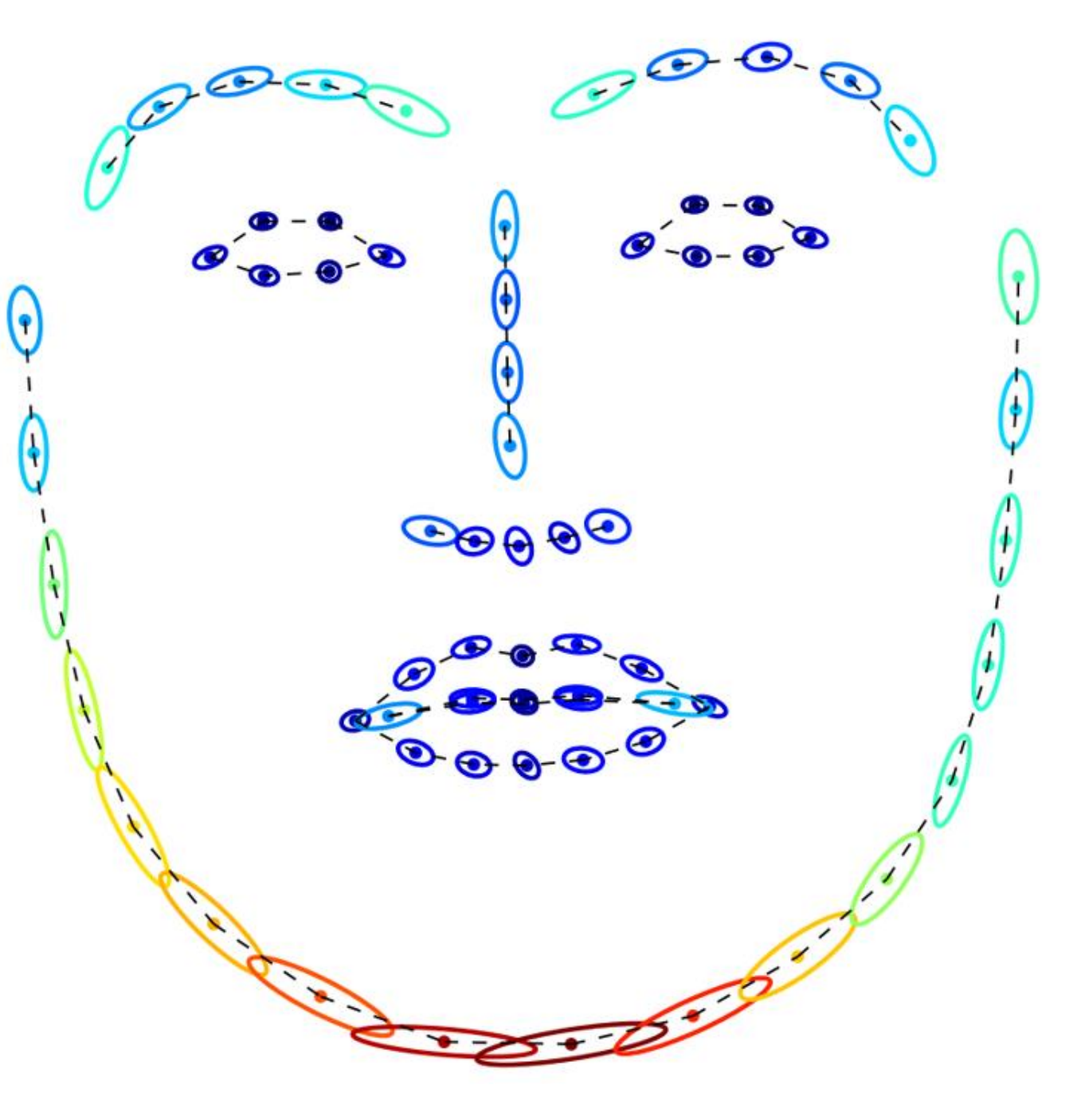}}
\subfigure[Baseline]
{\includegraphics[width=1.01in]{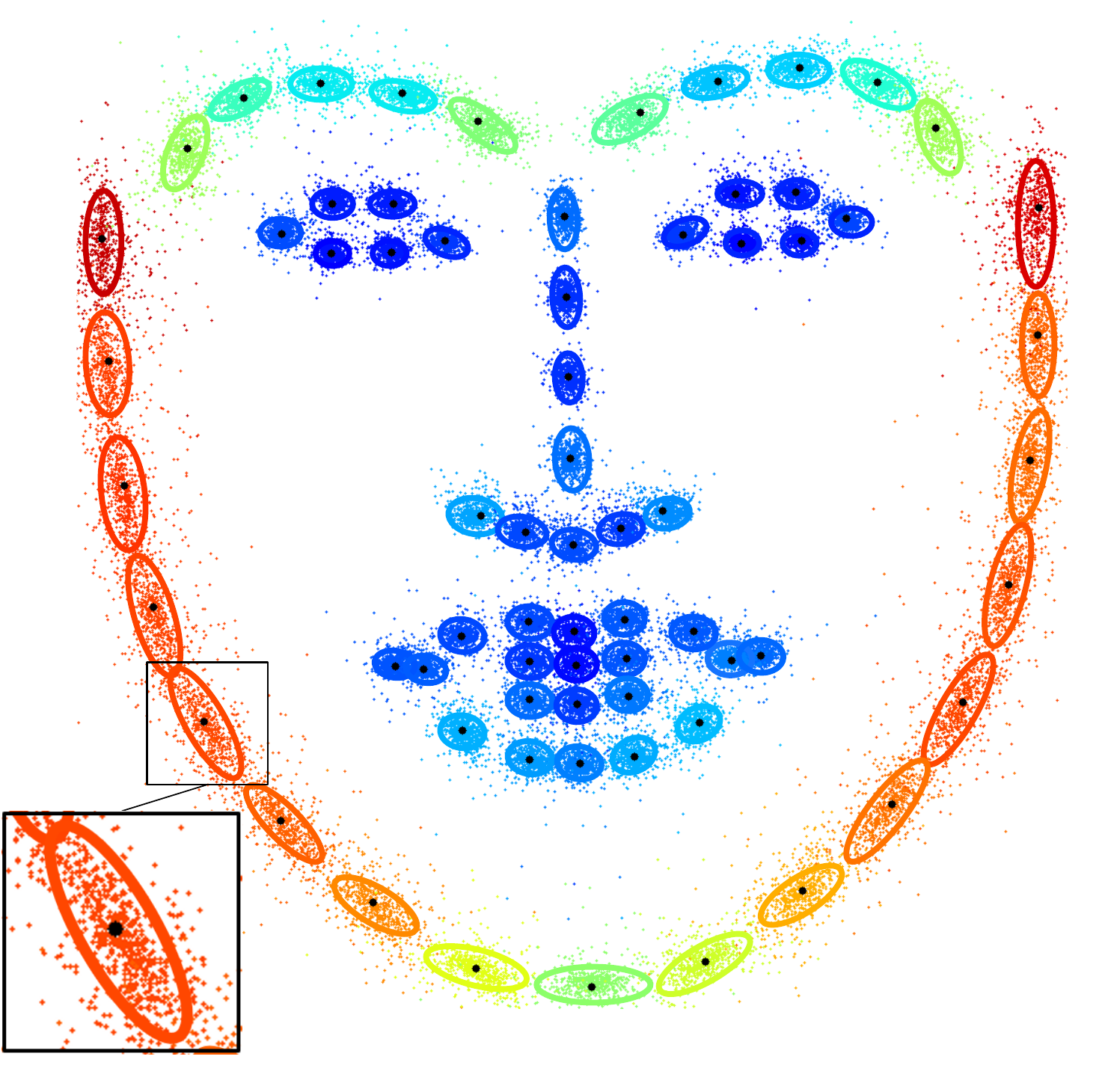}}
\subfigure[ADNet]
{\includegraphics[width=1.26in]{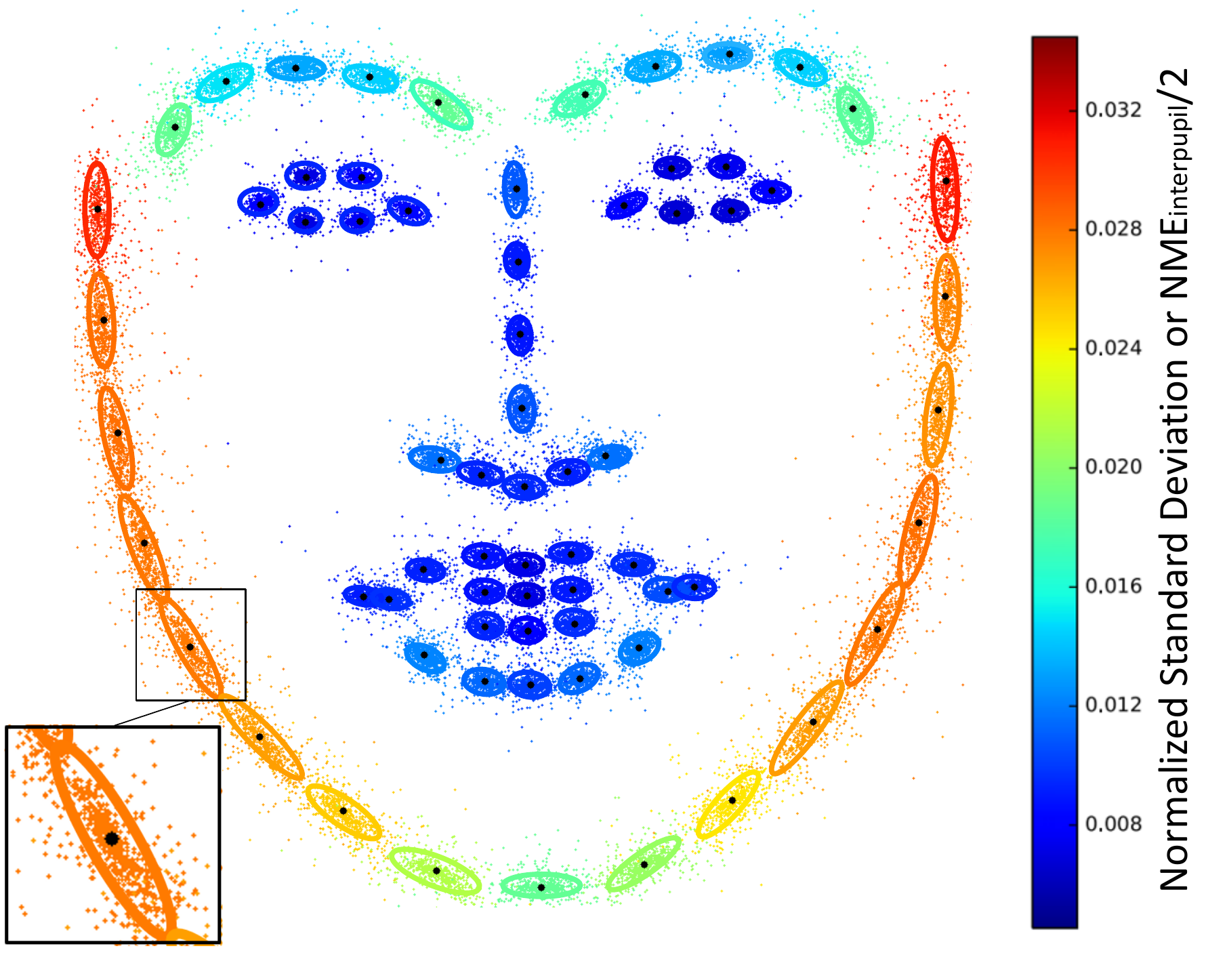}}
   \caption{Manual annotations variance v.s. Prediction error distribution on 300W dataset.
   (a) The variance of the manual annotations for each landmarks cited from 300W \cite{sagonas2016300}, the ellipses is colored by standard deviation normalized by face size. (b) The prediction error distribution of baseline model, each point represent the relevant position between predicted landmark and its corresponding ground truth, colored by the average NME$_{inter pupil}$ / 2 of that landmark, in general face size is twice of the inter pupil. (c) The prediction error distribution of ADNet model.}
\label{figure:error_distribution}
\end{figure}

In our research, error-bias is regarded as the nature of error direction and proves to be highly conducive for model understanding, thus deserving a thorough investigation to fill in the research gap. 

Figure~\ref{figure:error_distribution}.(a) demonstrates the labeling-bias of landmarks location on 300W dataset, deduced by the anisotropic standard deviation of each point, especially the points located on the boundary. By the observation, we trained a common face alignment model with 300W dataset, whose setting is the same as the baseline model described at the end of Section~\ref{subsection:evaluation_metrics}, and rendered the error distribution on the whole test dataset in Figure~\ref{figure:error_distribution}.(b), in which the error means the relative offset from predicted position to ground truth. It is found that the error-bias exists on common face alignment model and reaches 33\% relative rate, which is highly consistent with labeling-bias in Figure~\ref{figure:error_distribution}.(a). Based on the finding, we speculate that model easier converges the error in normal than in tangent direction because of noisy label and semantic confusion in tangent direction. Inspired by this, in order to test the applicability, stronger and weaker constraint are respectively imposed in normal and tangent direction by the proposed ADNet instead of isotropic loss. Similar to Figure~\ref{figure:error_distribution}.(b), the corresponding error distribution is shown in Figure~\ref{figure:error_distribution}.(c), in which the error-bias becomes higher at 48\%, but the error distribution becomes compact. To conclude, these three figures support our guideline that imposing stronger constraint to normal than to tangent.

In order to improve the localization capability of face alignment and fully leverage the error-bias, this paper proposes an end-to-end training framework involving devised Symmetric Direction Loss and Anisotropic Attention Module. Specifically, Anisotropic Direction Loss disentangles the landmark errors into normal error and tangent error, and imposes strong constraint in normal error and weak constraint in tangent error for coordinate regression, which is an improved $L_n$ loss. Anisotropic Attention Module combines point heatmap and edge heatmap into one heatmap, which contains both landmarks information and local boundary information. Applying the combined attention heatmap as a mask to the landmarks heatmap, the model has high tolerance to tangent direction and low tolerance to normal direction. These two modules are highly aligned with the proposed guideline. Some previous works, such as LAB~\cite{wu2018look}, PropNet~\cite{huang2020propagationnet}, incorporates boundary information into CNN by attention and can be treated as special cases for the proposed guideline. ADNet enables both heatmap regression and coordinate regression optimization, and the anisotropic attention mask in it acts like a gabor filter supervised by both point and edge information. 

The method is evaluated on several academic datasets, 300W \cite{sagonas2013300}, WFLW \cite{wu2018look} and COFW \cite{burgos2013robust}. All of them achieve the start-of-the-art performance, which demonstrates the effectiveness and robustness of the method.

In summary, the main contributions of this paper is as follows:
\begin{itemize}
  \item Unveiling error-bias on error directions of face alignment, which is highly consistent with labeling-bias by human, strong in tangent direction and weak in normal direction, and based on this firstly proposing the guideline to leverage error-bias in face alignment by magnification instead of suppression.
  \item Devising Anisotropic Direction Loss to assign uneven loss weights to the disentangled normal and tangent error for each landmarks coordinate and magnify the error-bias by the proposed guideline.
  \item Proposing Anisotropic Attention Module to generate anisotropic attention mask for each landmark heatmap and magnify the error-bias again by the proposed guideline.
  \item Constructing an advanced end-to-end training pipeline and implementing extensive experiments on various datasets, the result outperforms other state-of-the-art methods.
\end{itemize}

\section{Related Work}
\label{section:related_work}

In the course of face alignment development, some classic approaches of face alignment were proposed in the 1990s, e.g. AAM ~\cite{cootes2001active,saragih2007nonlinear,sauer2011accurate,matthews2004active,kahraman2007active}, ASM~ \cite{cootes1992active,cootes1995active,milborrow2008locating} and cascade regression~ \cite{feng2015cascaded,xiong2013supervised}. In recent years, driven by vigorous development of Deep Convolutional Neural Network (DCNN), CNN-based face alignment methods have achieved state-of-the-art performance and have drawn close attention from researchers. Currently, the spotlight is centered on two main branches of face alignment, that are coordinate regression and heatmap regression.

\noindent\textbf{Coordinate regression} methods \cite{sun2013deep,toshev2014deeppose,trigeorgis2016mnemonic,lv2017deep,zhang2014coarse,zhou2013extensive} directly regress facial landmarks based on the input without postprocessing. To solve this problem, Zhang \cite{zhang2015learning} involves multiple tasks, learning landmarks and facial attributes into the model concurrently. Then MDM \cite{trigeorgis2016mnemonic} designs a pipeline to train the model from coarse to fine by focusing more on detailed local information. The basic combination, ResNet \cite{he2016deep}, DenseNet \cite{huang2017densely} with L1, L2, Smooth L1 or wing loss \cite{feng2018wing} are commonly used in this type of methods.

\noindent\textbf{Heatmap regression} methods \cite{wei2016convolutional,dong2018style,deng2019joint,newell2016stacked,bulat2016convolutional} predict an intermediate heatmap for each landmark, and then the highest response point of each heatmap or near it is the final detected coordinate of landmark, where UNet \cite{ronneberger2015u} and stacked HG \cite{newell2016stacked} are applied frequently. Adaptive wing loss \cite{wang2019adaptive} and focal wing loss \cite{huang2020propagationnet} are proposed to balance the weight of easy sample and hard sample. Boundary information is introduced into face alignment by LAB \cite{wu2018look}, PropNet \cite{huang2020propagationnet} and ACENet \cite{huang2020ace}, which supplies more structure information and helps network to know which region should be paid more attention to. Usually, heatmap regression methods outperform coordinate regression methods with the effort of complex and tricky postprocessing.

Apart from the popular models, there are techniques that can further advance face alignment performance, such as CoordConv~\cite{liu2018intriguing}, Anti-aliased CNN~\cite{zhang2019making}, Multi-view CNN block~\cite{bulat2017binarized} and attention mechanism~\cite{wu2018look}. Among them, CoordConv proves to be instrumental for coordinate transformation problem in the vision task, such as object detection and generative modeling. Anti-aliased CNN with a special pooling layer, has advantages in translation invariance or ranging degrees of translation dependence. The multi-view CNN block is proven to be beneficial for landmark localization on account of multiple receptive fields and the various scale of images those fields bring about. Furthermore, the attention mechanism is able to guide a CNN to study valuable features and focus on salient regions, thus gaining great popularity among scholars nowadays. 

\section{ADNet}
\label{section:ADNet}

\begin{figure*}
\begin{center}
\includegraphics[width=6.6in]{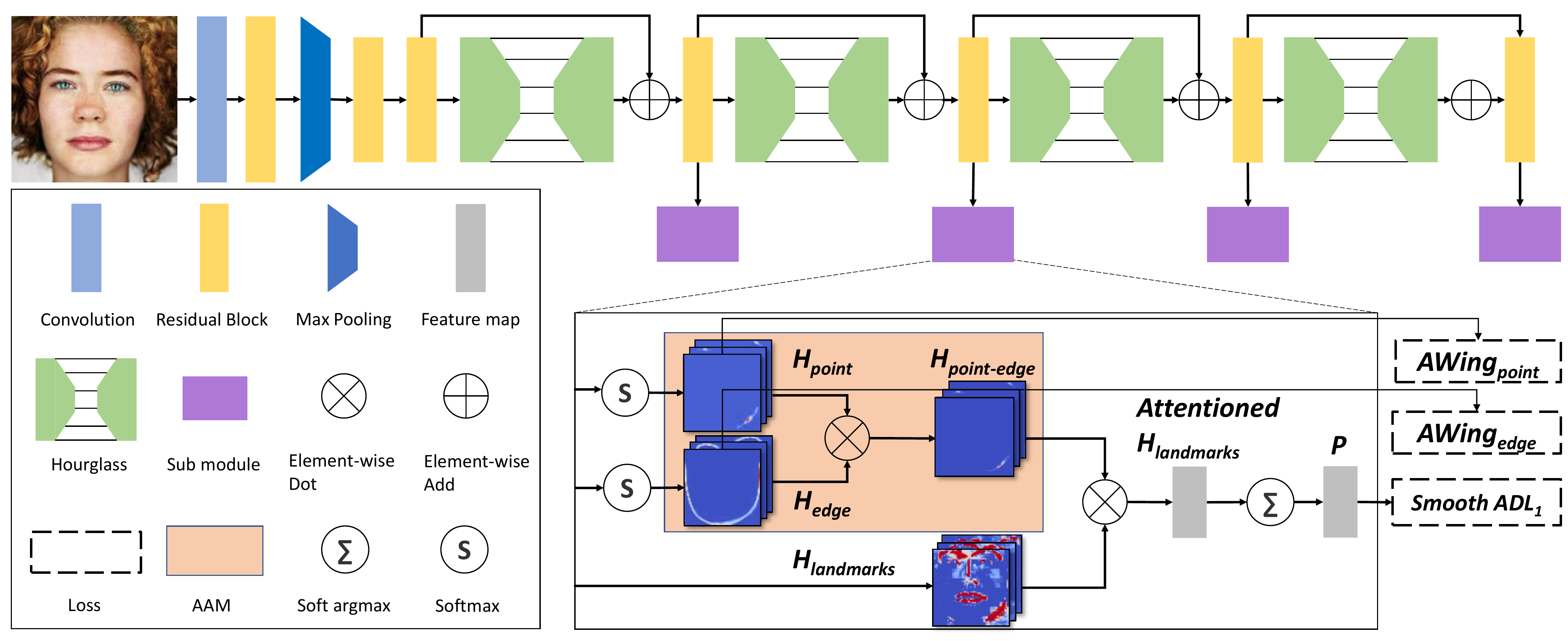}
\end{center}
   \caption{Overview of our ADL training framework.The backbone is constructed by stack four hourglass modules, and each hourglass module connected with three head branches, point attention, edge attention and final landmark regression branch.}
\label{figure:framework}
\end{figure*}

We design a network that employs stacked 4 hourglasses (HGs) \cite{bulat2017far} as backbone. In each HG structure, three heatmaps are generated, respectively corresponding to a \emph{landmarks heatmap}, an \emph{edge heatmap} and a \emph{point heatmap}. All heatmaps contribute losses to model training in different ways.
Based on these heatmaps, an \emph{anisotropic attention mask} is generated from the point and edge heatmaps. The attention mask can then impose anisotropic supervision upon the landmarks training, where an anisotropic direction loss is applied to the predicted landmark coordinates that are formed through soft-argmax operation. We also leverage components such as coordconv \cite{liu2018intriguing} and anti-aliased blocks \cite{zhang2019making} to improve performance further, as is also proposed by \cite{huang2020propagationnet}.
We name this network the ADNet, with AD standing for anisotropic direction. The overview structure of ADNet is detailedly illustrated in Figure~\ref{figure:framework}.

\subsection{Anisotropic Direction Loss (ADL)}

Before proposing the new loss, we would like to review the \emph{$L_n$} loss in Equation~\ref{equation:Ln}. Two of its special cases are \emph{$L_1$} and \emph{$L_2$}, which are commonly used in machine learning tasks, e.g. regression, detection and GAN.

\begin{equation}
L_{n}(p_i, \hat{p}_i) = |p_i-\hat{p}_i|^n
\label{equation:Ln}
\end{equation}

where $p_i$ and $\hat{p_i}$ are respectively the predicted value and ground truth of coordinate in $i$th landmarks. However, when utilizing \emph{$L1$} loss, model learns much noise instead of valid information from easy samples and when applying \emph{$L_2$} to outliers, model easily gets to gradient explosion. With the development of \emph{$L_n$}, {$Smooth\ L_1$} is proposed to solve these two problems by piecewise-defining as Equation~\ref{equation:smooth_L1}.

\begin{equation}
\textnormal{\emph{Smooth}}L_{1}(p_i, \hat{p}_i) =
\begin{cases}
\ 0.5 L_{2}(p_i, \hat{p}_i),& |p_i-\hat{p}_i| < 1 \\
\ L_{1}(p_i, \hat{p}_i) - 0.5,& otherwise \\
\end{cases}
\label{equation:smooth_L1}
\end{equation}

The \emph{$L_n$} series losses treat the error isotropically, even if there is error-bias on error direction of model or labeling-bias on annotations, such as in landmarks regression. In other words, they ignore the anisotropy of errors in face alignment and give the same weight to loss in all directions.

\begin{figure}
\begin{center}
\includegraphics[width=3in]{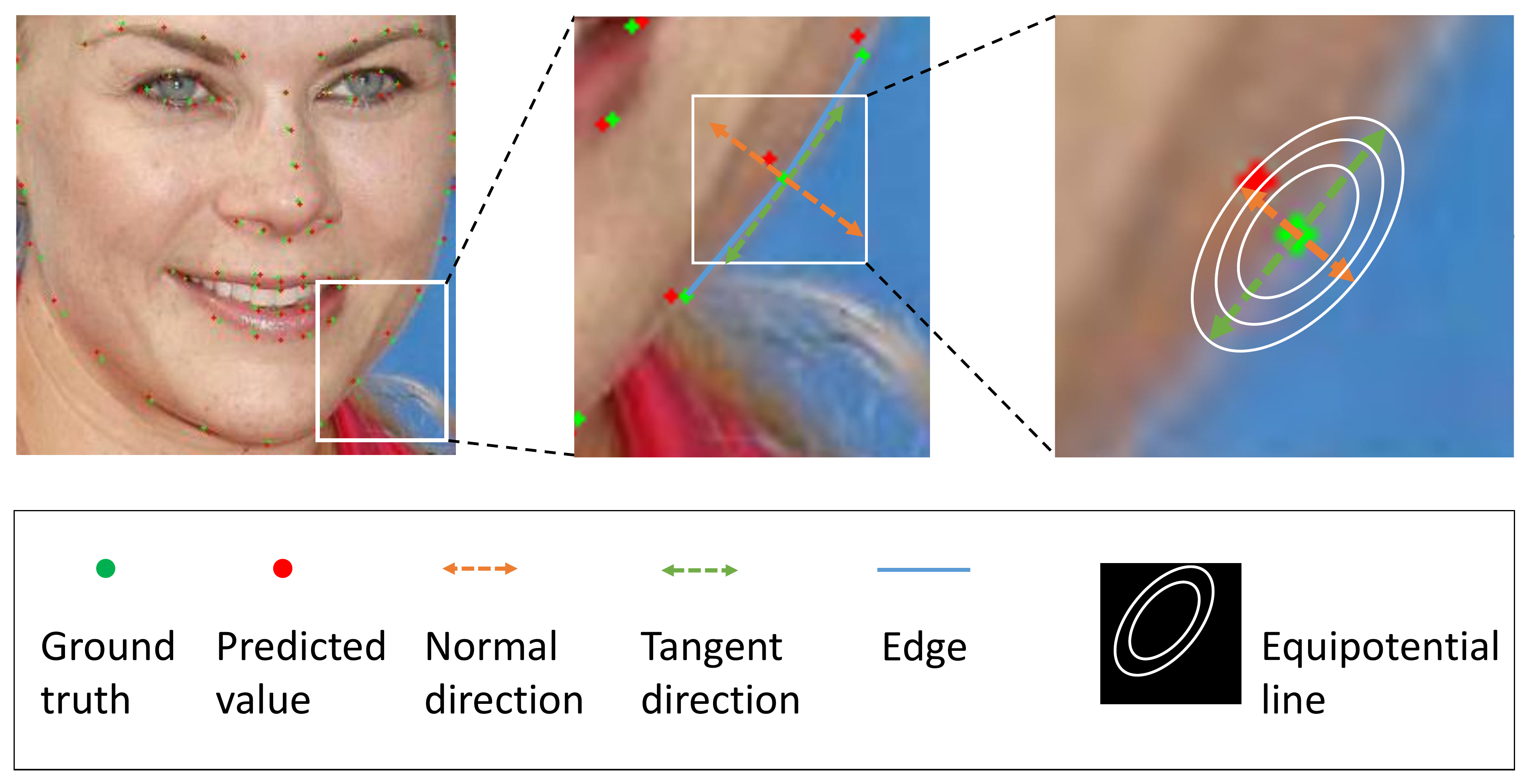}
\end{center}
   \caption{Diagram of Anisotropic Direction Loss (ADL). In the right figure, the green point ($\hat{p_i}$) attracts red point ($p_i$) under force filed, which is strong in normal direction and weak in tangent directions. According to the proposed guideline, the model studies more in normal direction.}
\label{figure:ADL}
\end{figure}

According to the guideline of imposing strong constraint in normal direction and weak constraint in tangent direction, we propose the anisotropic direction loss (ADL), shown in Figure~\ref{figure:ADL}, which disentangles the error into two mutually orthogonal directions, namely normal error and tangent error, and put anisotropic loss weight to them, defined as follows:

\begin{equation}
\begin{aligned}
\textnormal{\emph{ADL}}_{n}(p_i, \hat{p}_i) = (\frac{2\lambda}{1+\lambda}|N(p_i, \hat{p}_i) \cdot (p_i-\hat{p}_i)|^2 + \\ \frac{2}{1+\lambda}|T(p_i, \hat{p}_i) \cdot (p_i-\hat{p}_i)|^2)^\frac{n}{2}
\end{aligned}
\label{equation:ADLn}
\end{equation}

Where \emph{$N(p_i, \hat{p}_i)$} and \emph{$T(p_i, \hat{p}_i)$} are unit vectors to disentangle error into two sub-errors, when the \emph{$\hat{p}_i$} locates on the edge, they are unit normal and tangent vectors respectively, otherwise, they would be unit error vectors; and \emph{$\lambda$} refers to the hyper parameter to adjust constraint strength in these two sub-errors. Refer to the equation below for the corresponding {$Smooth\ ADL_{1}$} loss:

\begin{equation}
\begin{aligned}
& \hspace{-16pt} \textnormal{\emph{SmoothADL}}_{1}(p_i, \hat{p}_i) = \\
& \hspace{16pt} \begin{cases}
\ 0.5 \: \textnormal{\emph{ADL}}_{2}(p_i, \hat{p}_i),& |p_i-\hat{p}_i| < 1 \\
\ \textnormal{\emph{ADL}}_{1}(p_i, \hat{p}_i) - 0.5,& otherwise \\
\end{cases}
\end{aligned}
\label{equation:smooth_ADL1}
\end{equation}

The \emph{ADL$_n$} loss can be treated as a more general form of the \emph{$L_n$} loss. When \emph{$\lambda=1$} or \emph{$\hat{p}_i$} does not subordinate to any edge, \emph{ADL$_n$} exactly degenerates into \emph{$L_n$}.

Generally, most landmarks locate on edges of either face contour or five sense organs. Hence, for these landmarks, we get the normal direction by calculating the slope from its directly adjacent points on the edge, namely \emph{$\hat{p}_{pre_i}$} and \emph{$\hat{p}_{next_i}$}, which can be obtained from a pre-defined template of landmarks. And as for other landmarks not associating any edges, we still put isotropic constraint on all the directions, thus, both normal direction and tangent direction are equal to error direction. Based on this, normal and tangent direction can respectively be defined as below:

\begin{equation}
N(p_i, \hat{p}_i) =
\begin{cases}
\frac{\hat{p}_{pre_i} + \hat{p}_{next_i} - 2\hat{p}_i}{|\hat{p}_{pre_i} + \hat{p}_{next_i} - 2\hat{p}_i|},& \hat{p}_i\ in\ edge \\
\frac{p_i - \hat{p}_i,}{|p_i - \hat{p}_i|}& otherwise \\
\end{cases}
\label{equation:normal_direction}
\end{equation}

\begin{equation}
T(p_i, \hat{p}_i) =
\begin{cases}
S \times N(p_i, \hat{p}_i),& \hat{p}_i\ in\ edge \\
N(p_i, \hat{p}_i),& otherwise \\
\end{cases}
\label{equation:tangent_direction}
\end{equation}

where $S$ is a skew-symmetric matrix $\bigl(\begin{smallmatrix}
0&1 \\ -1&0
\end{smallmatrix} \bigr)$.

\subsection{Anisotropic Attention Module (AAM)}

As introduced in Section~\ref{section:related_work}, coordinate regression and heatmap regression are widely applied in landmarks detection. Inspired by the proposed guideline, ADL loss is devised to deal with coordinate regression task under the bias of error direction. And as for heatmap regression task, in order to enhance the localization capability on normal direction, anisotropic attention module (AAM) is designed through supervising the network to focus more on tangent direction. In previous boundary-aware methods, the attention mask only contains boundary information, thus merely from which it is incapable of getting any point information.

To leverage the complete semantic information offered by prior information, the points and its directly connected edges, anisotropic attention module outputs the \emph{anisotropic heatmap} by combining \emph{point heatmap} and \emph{edge heatmap} in Equation~\ref{equation:attention_heatmap_fusion}, named \emph{point-edge heatmap}. When utilizing the \emph{point-edge heatmap} as attention mask, \textbf{\emph{the local region of target point with its partial edge has high response}}, which means it contains both edge and point information and can guide model training in normal and tangent anisotropically.
Therefore, using \emph{soft argmax} operation with the ability of mapping heatmap to coordinate, model accumulates landmarks coordinates mainly on the local region of attentioned landmarks heatmap with tangent information instead of on whole region of the landmarks heatmap.

\begin{equation}
H_{point-edge} = H_{point} \otimes H_{edge}
\label{equation:attention_heatmap_fusion}
\end{equation}

Where \emph{$H_{point}$} is the \emph{point heatmap} and \emph{$H_{edge}$} is the \emph{edge heatmap}. Both of them are generated from the feature map of each HG and are supervised on two ground truth concurrently, point and edge heatmaps, by Awing \cite{wang2019adaptive} loss in Equation~\ref{equation:awing_loss}. Then the combined $H_{point-edge}$ serves as attention mask in landmarks coordinates regression by $Smooth\ ADL_{1}$ loss in Equation~\ref{equation:smooth_ADL1}. In this process, it is interesting that the shape of salient region in \emph{point heatmap $H_{point}$} is more like a rough edge than a point region in gaussian distribution. This is caused by the supervisory signal from $Smooth\ ADL_{1}$ loss and is conducive to generate \emph{point-edge heatmap} with strong tangent information. 

\begin{equation}
\textnormal{\emph{AWing}}(y, \hat{y}) =
\begin{cases}
\ \omega \ln(1+|\frac{y-\hat{y}}{\epsilon}|^{\alpha-y}) ,& |y-\hat{y}| < \theta \\
\ A|y-\hat{y}|-C,& otherwise \\
\end{cases}
\label{equation:awing_loss}
\end{equation}

The detailed description and default values of hyper parameters could refer to AWing \cite{wang2019adaptive} loss, from which both \emph{$H_{point}$} and \emph{$H_{edge}$} adopt the same setting to learn the ground truth of respective heatmaps.

\begin{equation}
P = \textnormal{\emph{soft argmax}}(H_{landmarks} \otimes H_{point-edge})
\label{equation:predicted_coordinate}
\end{equation}

Where \emph{$H_{landmarks}$} is also one of the feature maps from each HG, and \emph{$P$} is the predicted landmarks value by applying soft argmax operation to \emph{$H_{landmarks}$}. Then network could learn \emph{$P$} by annotated $\hat{P}$ for each training sample under the constraint of $Smooth\ ADL_{1}$.

Finally, the holistic loss of our architecture is:

\begin{equation}
\begin{aligned}
L_{\textnormal{\emph{ADNet}}} =& \: \textnormal{\emph{SmoothADL}}_{1} \\ 
&+ \alpha \cdot \textnormal{\emph{AWing}}_{edge} + \beta \cdot \textnormal{\emph{AWing}}_{point}
\end{aligned}
\label{equation:overall_loss}
\end{equation}

Where \emph{$\alpha$} and \emph{$\beta$} are the loss weights to balance different tasks, we empirically set both as 10 in the paper.

\section{Experiment}
\label{section:experiment}

\subsection{Implementation Details}
\label{subsection:implementation_details}

To generate the inputs of our model, we cropped face regions and resized them into $256 \mathsf{\times} 256$.
We applied augmentation techniques to the training data by random rotation (18\degree), random scaling ($\pm10\%$), random crop ($\pm5\%$), random gray (20\%), random blur (30\%), random occlusion (40\%) and random horizontal flip (50\%).
Regarding the architecture, we adopt four stacked hourglass modules. 
Each hourglass outputs three $64 \times 64$ feature maps (landmark, edge, and point heatmaps).
To train our model, we empirically set the hyper-parameters of Gaussian sigma as 1.5 in point heatmap generation, line width as 1.0 in edge heatmap generation, and $\lambda=2.0$ in the anisotropic direction loss function. 
We employed Adam optimizer with the initial learning rate of $1\times10^{-3}$ and reduced the learning rate by 1/10 at each epoch of 200, 350, and 450.
The model was trained on four GPUs (16GB NVIDIA Tesla P100), where the batch size of each GPU is 8.

\subsection{Evaluation Metrics}
\label{subsection:evaluation_metrics}

\noindent\textbf{Normalized Mean Error (NME)} is a widely-used standard metric to evaluate landmark accuracy for face alignment algorithms, which is defined as

\begin{equation}
\textnormal{NME}(P, \hat{P}) = \frac{1}{N_P}\sum\limits_{i=1}^{N_P}\frac{\|p_i-\hat{p}_i\|_2}{d}
\label{equation:NME}
\end{equation}

Where $P$ and $\hat{P}$ denote the predicted and ground-truth coordinates of landmarks, respectively, $N_P$ is the number of landmarks, and $d$ is the reference distance to normalize the absolute errors.
Usually, $d$ could be inter-ocular (distance between outer eye corners) or inter-pupils (distance between pupil centers) distance. This paper uses inter-ocular distance as the normalization factor for 300W \cite{sagonas2013300}, WFLW \cite{wu2018look}, and inter-pupils for 300W \cite{sagonas2013300} and COFW \cite{burgos2013robust}.

\vspace{5pt}
\noindent\textbf{Failure Rate (FR)} is a metric to evaluate the robustness of algorithms in terms of NME. 
Samples having larger NME than a pre-defined threshold are regarded as failed prediction.
FR is defined by the percentage of failed examples over the whole dataset.
We set the thresholds by 5\% for 300W \cite{sagonas2013300}, and 10\% for COFW \cite{burgos2013robust} and WLFW \cite{wu2018look}.

\vspace{5pt}
\noindent\textbf{Area Under Curve (AUC)} is another widely-adopted metric for face alignment task.
It can be calculated by using Cumulative Error Distribution (CED) curve.
The horizontal axis of CED plot indicates the target NME (ranging between 0 and the pre-defined threshold), and the CED value at the specific point means the rate of samples whose NME are smaller than the specific target NME.

\subsection{Method Comparison}
\label{subsection:method_comparison}

To compare the performance of ADNet with the state-of-the-art competitors, we conduct evaluation on three public datasets: COFW \cite{burgos2013robust}, 300W \cite{sagonas2013300} and WFLW \cite{wu2018look}.

\vspace{5pt}
\noindent\textbf{COFW} \cite{burgos2013robust} contains a wide range of head poses and heavy occlusions, with 1,345 training images and 507 testing images. Each face has 29 annotated landmarks. 

The experimental results on the COFW dataset are shown in Table~\ref{table:COFW}, where all the models were tested under the same condition without any additional annotations.
As tabulated in the table, our method outperforms the competitors by a wide margin in NME and FR$_{10\%}$.
Compared with the second leading method, ADNet decreases NME by 5.3\% in NME, which implies that the model is robust to large head poses and heavy occlusion despite the sparse landmarks definition.

\begin{table}
\begin{center}
\begin{tabular}{|l|c|c|c|}
\hline
Method & NME & FR$_{10\%}$ & AUC$_{10\%}$ \\
\hline
Human \cite{burgos2013robust} & 5.60 & - & - \\
RCPR \cite{burgos2013robust} & 8.50 & 20.00 & - \\
TCDCN \cite{zhang2014facial} & 8.05 & - & - \\
DAC-CSR \cite{feng2017dynamic} & 6.03 & 4.73 & - \\
Wu et al \cite{wu2015robust} & 5.93 & - & - \\
Wing \cite{feng2018wing} & 5.44 & 3.75 & - \\
DCFE \cite{valle2018deeply} & 5.27 & 7.29 & 0.3586 \\
Awing \cite{wang2019adaptive} & 4.94 & 0.99 & \textbf{0.6440} \\
\hline
\textbf{ADNet(Ours)} & \textbf{4.68} & \textbf{0.59} & 0.5317 \\
\hline
\end{tabular}
\end{center}
\caption{Comparing with state-of-the-art methods on COFW.}
\label{table:COFW}
\end{table}

\vspace{5pt}
\noindent\textbf{300W} \cite{sagonas2013300} is a widely-adopted dataset for face alignment, with 3,148 images for training and 689 images for testing. The testing data is divided into two sub-categories: 554 for the common subset, and 135 for the challenging subset. All the data are manually labelled with 68 landmarks.

Table~\ref{table:300W} compares our methods with the other approaches using different normalization factors: inter-ocular and inter-pupils distances.
In both settings, our method yields the competitive accuracy, usually one of the best two.
In particular, for inter-ocular normalization, ADNet improves the metric by 4.6\% and 7.0\% in NME over the previous SOTA method on the fullset and the common subset, respectively.
As shown in Table~\ref{table:300W}, ADNet achieves a larger improvement on the common subset than the challenging subset, which could be due to the inaccurate labels in the challenging subset. For instance, in the challenging subset of 300W, we observed some samples having inaccurate landmark labels on the chin regions due to wrong bounding boxes, which could lead to the domain gap issue and degraded performance.

\begin{table}
\begin{center}
\begin{tabular}{|l|c|c|c|}
\hline
Method & \makecell{Common \\ Subset} & \makecell{Challenging \\ Subset} & \makecell{Fullset} \\
\hline
\multicolumn{4}{|c|}{Inter-pupil Normalization} \\
\hline
CFAN \cite{zhang2014coarse} & 5.50 & 16.78 & 7.69 \\
SDM \cite{xiong2013supervised} & 5.57 & 15.40 & 7.50 \\
LBF \cite{ren2014face} & 4.95 & 11.98 & 6.32 \\
CFSS \cite{zhu2015face} & 4.73 & 9.98 & 5.76 \\
TCDCN \cite{zhang2015learning} & 4.80 & 8.60 & 5.54 \\
MDM \cite{trigeorgis2016mnemonic} & 4.83 & 10.14 & 5.88 \\
RAR \cite{xiao2016robust} & 4.12 & 8.35 & 4.94 \\
DVLN \cite{wu2017leveraging} & 3.94 & 7.62 & 4.66 \\
TSR \cite{lv2017deep} & 4.36 & 7.56 & 4.99 \\
DSRN \cite{miao2018direct} & 4.12 & 9.68 & 5.21 \\
RCN$_+$(L+ELT) \cite{honari2018improving} & 4.20 & 7.78 & 4.90 \\
DCFE \cite{valle2018deeply} & 3.83 & 7.54 & 4.55 \\
LAB \cite{wu2018look} & 3.42 & 6.98 & 4.12 \\
Wing \cite{feng2018wing} & \textbf{3.27} & 7.18 & \textbf{4.04} \\
AWing \cite{wang2019adaptive} & 3.77 & 6.52 & 4.31 \\
\hline
\textbf{ADNet(Ours)} & 3.51 & \textbf{6.47} & 4.08 \\
\hline
\multicolumn{4}{|c|}{Inter-ocular Normalisation} \\
\hline
PCD-CNN \cite{kumar2018disentangling} & 3.67 & 7.62 & 4.44 \\
CPM+SBR \cite{dong2018style} & 3.28 & 7.58 & 4.10 \\
SAN \cite{dong2018style} & 3.34 & 6.60 & 3.98 \\
LAB \cite{wu2018look} & 2.98 & 5.19 & 3.49 \\
DeCaFA \cite{dapogny2019decafa} & 2.93 & 5.26 & 3.39 \\
DU-Net \cite{tang2018quantized} & 2.90 & 5.15 & 3.35 \\
LUVLi \cite{kumar2020luvli} & 2.76 & 5.16 & 3.23 \\
AWing \cite{wang2019adaptive} & 2.72 & \textbf{4.52} & 3.07 \\
\hline
\textbf{ADNet(Ours)} & \textbf{2.53} & 4.58 & \textbf{2.93} \\
\hline
\end{tabular}
\end{center}
\caption{Comparing with state-of-the-art methods on 300W.}
\label{table:300W}
\end{table}

\vspace{5pt}
\noindent\textbf{WFLW} \cite{wu2018look} is a challenging dataset constructed from in-the-wild face images. It also provides rich attribute labels such as occlusion, pose, make-up, illumination, blur and expression. And it consists of 7,500 faces for training and 2,500 faces for testing, with 98 annotated landmarks.

Using the WFLW dataset, we aim to evaluate the models in various specific scenarios by making use of the annotated labels such as pose, expression, illumination and occlusion. In addition, in order to demonstrate the stability of landmark localization, we employ three metrics: NME, FR (10\%) and AUC (10\%).
As tabulated in Table~\ref{table:WFLW}, ADNet outperforms the other state-of-the-art methods significantly in most of subsets.

\begin{table*}
\begin{center}
\begin{tabular}{|l|c|c|c|c|c|c|c|c|}
\hline
Metric & Method & Testset & \makecell{Pose \\ Subset} & \makecell{Expression \\ Subset} & \makecell{Illumination \\ Subset} & \makecell{Make-up \\ Subset} & \makecell{Occlusion \\ Subset} & \makecell{Blur \\ Subset} \\
\hline
\multirow{6}*{NME(\%)} & ESR \cite{cao2014face} & 11.13 & 25.88 & 11.47 & 10.49 & 11.05 & 13.75 & 12.20 \\
 & SDM \cite{xiong2013supervised} & 10.29 & 24.10 & 11.45 & 9.32 & 9.38 & 13.03 & 11.28 \\
 & CFSS \cite{zhu2015face} & 9.07 & 21.36 & 10.09 & 8.30 & 8.74 & 11.76 & 9.96 \\
 & DVLN \cite{wu2017leveraging} & 6.08 & 11.54 & 6.78 & 5.73 & 5.98 & 7.33 & 6.88 \\
 & LAB \cite{wu2018look} & 5.27 & 10.24 & 5.51 & 5.23 & 5.15 & 6.79 & 6.12 \\
 & Wing \cite{feng2018wing} & 5.11 & 8.75 & 5.36 & 4.93 & 5.41 & 6.37 & 5.81 \\
 & DeCaFA \cite{dapogny2019decafa} & 4.62 & 8.11 & 4.65 & 4.41 & 4.63 & 5.74 & 5.38 \\
 & AWing \cite{wang2019adaptive} & 4.36 & 7.38 & 4.58 & 4.32 & 4.27 & 5.19 & 4.96 \\
 & LUVLi \cite{kumar2020luvli} & 4.37 & - & - & - & - & - & - \\
 & \textbf{ADNet(Ours)} & \textbf{4.14} & \textbf{6.96} & \textbf{4.38} & \textbf{4.09} & \textbf{4.05} & \textbf{5.06} & \textbf{4.79} \\
\cline{2-9}
 & PropNet & 4.05 & 6.92 & \textbf{3.87} & 4.07 & 3.76 & 4.58 & 4.36 \\
 & \textbf{ADNet$^*$(Ours)} & \textbf{3.98} & \textbf{6.56} & 4.02 & \textbf{3.87} & \textbf{3.62} & \textbf{4.36} & \textbf{4.21} \\
\hline
\multirow{6}*{FR$_{10\%}$} & ESR \cite{cao2014face} & 35.24 & 90.18 & 42.04 & 30.80 & 38.84 & 47.28 & 41.40 \\
 & SDM \cite{xiong2013supervised} & 29.40 & 84.36 & 33.44 & 26.22 & 27.67 & 41.85 & 35.32 \\
 & CFSS \cite{zhu2015face} & 20.56 & 66.26 & 23.25 & 17.34 & 21.84 & 32.88 & 23.67 \\
 & DVLN \cite{wu2017leveraging} & 10.84 & 46.93 & 11.15 & 7.31 & 11.65 & 16.30 & 13.71 \\
 & LAB \cite{wu2018look} & 7.56 & 28.83 & 6.37 & 6.73 & 7.77 & 13.72 & 10.74 \\
 & Wing \cite{feng2018wing} & 6.00 & 22.70 & 4.78 & 4.30 & 7.77 & 12.50 & 7.76 \\
 & DeCaFA \cite{dapogny2019decafa} & 4.84 & 21.40 & 3.73 & 3.22 & 6.15 & 9.26 & 6.61 \\
 & AWing \cite{wang2019adaptive} & 2.84 & 13.50 & 2.23 & 2.58 & 2.91 & 5.98 & 3.75 \\
 & LUVLi \cite{kumar2020luvli} & 3.12 & - & - & - & - & - & - \\
 & \textbf{ADNet(Ours)} & \textbf{2.72} & \textbf{12.72} & \textbf{2.15} & \textbf{2.44} & \textbf{1.94} & \textbf{5.79} & \textbf{3.54} \\
\cline{2-9}
 & PropNet & 2.96 & 12.58 & 2.55 & 2.44 & 1.46 & 5.16 & 3.75 \\
 & \textbf{ADNet$^*$(Ours)} & \textbf{2.00} & \textbf{9.20} & \textbf{1.59} & \textbf{1.72} & \textbf{1.26} & \textbf{4.48} & \textbf{2.59} \\
\hline
\multirow{6}*{AUC$_{10\%}$} & ESR \cite{cao2014face} & 0.2774 & 0.0177 & 0.1981 & 0.2953 & 0.2485 & 0.1946 & 0.2204 \\
 & SDM \cite{xiong2013supervised} & 0.3002 & 0.0226 & 0.2293 & 0.3237 & 0.3125 & 0.2060 & 0.2398 \\
 & CFSS \cite{zhu2015face} & 0.3659 & 0.0632 & 0.3157 & 0.3854 & 0.3691 & 0.2688 & 0.3037 \\
 & DVLN \cite{wu2017leveraging} & 0.4551 & 0.1474 & 0.3889 & 0.4743 & 0.4494 & 0.3794 & 0.3973 \\
 & LAB \cite{wu2018look} & 0.5323 & 0.2345 & 0.4951 & 0.5433 & 0.5394 & 0.4490 & 0.4630 \\
 & Wing \cite{feng2018wing} & 0.5504 & 0.3100 & 0.4959 & 0.5408 & 0.5582 & 0.4885 & 0.4918 \\
 & DeCaFA \cite{dapogny2019decafa} & 0.5630 & 0.2920 & \textbf{0.5460} & 0.5790 & 0.5750 & 0.4850 & 0.4940 \\
 & AWing \cite{wang2019adaptive} & 0.5719 & 0.3120 & 0.5149 & 0.5777 & 0.5715 & 0.5022 & 0.5120 \\
 & LUVLi \cite{kumar2020luvli} & 0.5770 & - & - & - & - & - & - \\
 & \textbf{ADNet(Ours)} & \textbf{0.6022} & \textbf{0.3441} & 0.5234 & \textbf{0.5805} & \textbf{0.6007} & \textbf{0.5295} & \textbf{0.5480} \\
\cline{2-9}
 & PropNet & 0.6158 & 0.3823 & \textbf{0.6281} & 0.6164 & 0.6389 & 0.5721 & 0.5836 \\
 & \textbf{ADNet$^*$(Ours)} & \textbf{0.6250} & \textbf{0.4036} & 0.6014 & \textbf{0.6295} & \textbf{0.6406} & \textbf{0.5896} & \textbf{0.5903} \\
\hline
\end{tabular}
\end{center}
\caption{Comparing with state-of-the-art methods on WFLW testing set. PropNet and ADNet$^*$(Ours) employ focal wing loss~\cite{huang2020propagationnet} by using the attribute labels provided by WFLW.}
\label{table:WFLW}
\end{table*}

\begin{figure}
\begin{center}
\includegraphics[width=3.2in]{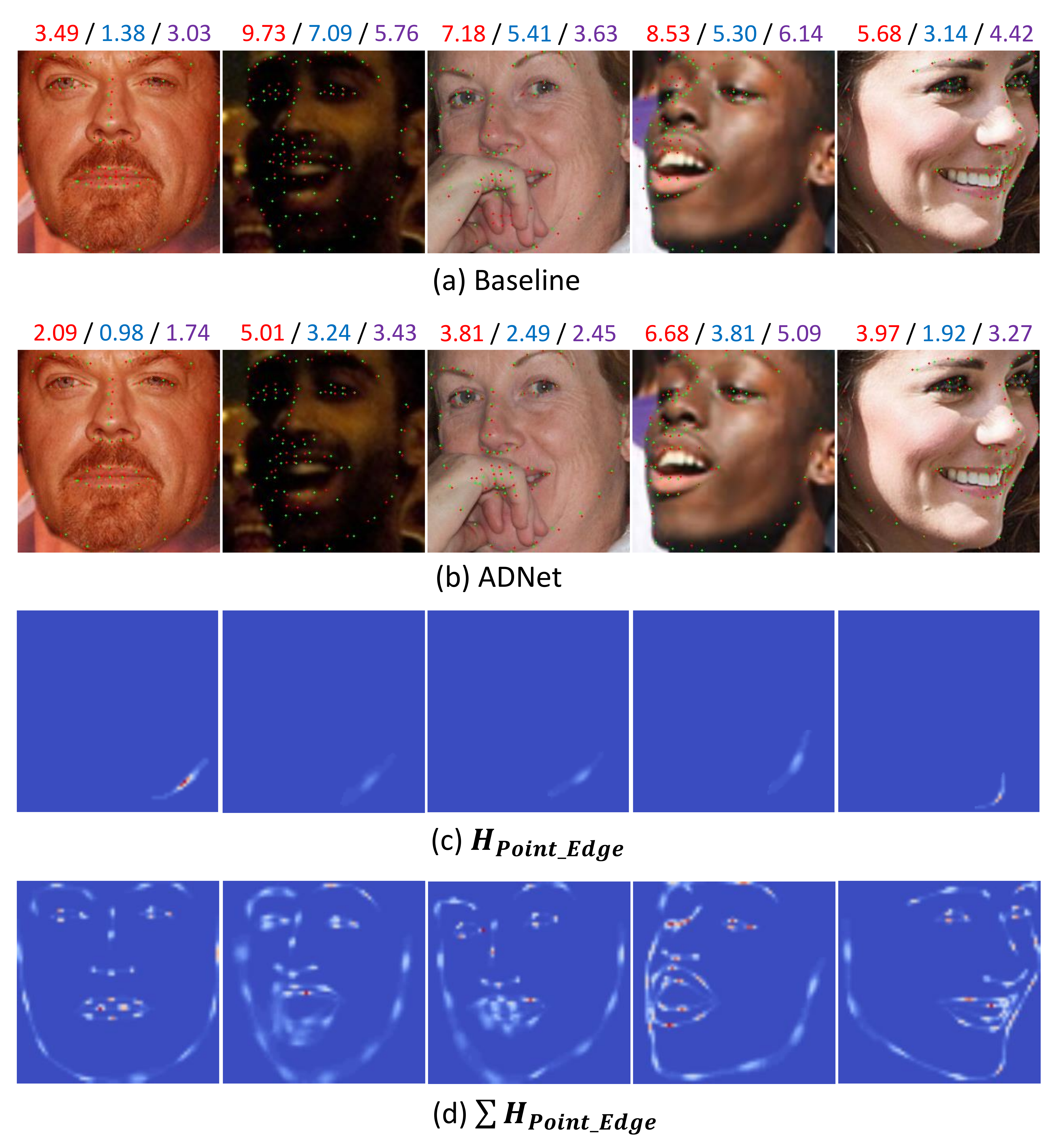}
\end{center}
   \caption{Samples from 300W fullset. (\textcolor[rgb]{1, 0, 0}{Red} indicates NME, \textcolor[rgb]{0, 0.44, 0.75}{blue} indicates normal NME, and \textcolor[rgb]{0.44, 0.19, 0.63}{purple} indicates tangent NME.) (a) is the baseline result without ADL and AAM, (b) is the result of ADNet, (c) is the output of a sample from \textbf{$H_{point-edge}$} and (d) is the accumulated output of half \textbf{$H_{point-edge}$} set for better visualization.}
\label{figure:visualization}
\end{figure}

\subsection{Ablation study}
\label{subsection:ablation_study}
To explore the efficacy and the contribution of each component of our proposed model, we conduct comprehensive ablation experiments.

\vspace{5pt}
\noindent\textbf{Evaluation on different $\lambda$ for ADL.}
To find the optimal hyper-parameters for ADL, we trained our model with different $\lambda$ on the 300W dataset. And AAM is not included so as to avoid 
ambiguity.
Instead of seeking the entire parameter space, we choose a few candidates of $\lambda$ as shown in Table~\ref{table:loss}.
It appears that, compared with the model without ADL ($\lambda=1.0$), the optimal model gained about 4\% improvement in NME when $\lambda=2.0$.
In addition, when $\lambda$ is set to 0.5 (more weight for the tangent direction), the alignment accuracy dropped, which aligns with our assumption about error bias in face alignment.
However, when $\lambda \geq 2.0$, the performance is not sensitive to the value of $\lambda$.
In our experiments, we chose $\lambda=2$ as the default setting.

\begin{table}[t]
\begin{center}
\begin{tabular}{|l|c|c|c|}
\hline
Method & NME(\%) & FR$_{5\%}$ & AUC$_{5\%}$ \\
\hline
$\lambda$=0.5 & 3.43 & 12.01 & 0.3504 \\
\hline
$\lambda$=1.0(w/o ADL) & 3.38 & 11.72 & 0.3653 \\
\hline
$\lambda$=2.0 & \textbf{3.23} & \textbf{10.45} & 0.4006 \\
\hline
$\lambda$=4.0 & 3.24 & 11.03 & \textbf{0.4027} \\
\hline
$\lambda$=8.0 & 3.27 & 11.47 & 0.3950 \\
\hline
\end{tabular}
\end{center}
\caption{The contribution of ADL under different $\lambda$. AAM is not used in these settings.
}
\label{table:loss}
\vspace{-2.0em}
\end{table}

\vspace{5pt}
\noindent\textbf{Evaluation on different attention modules for AAM.}
AAM plays a vital role in ADNet, which contributes the largest improvement in alignment accuracy.
In Table~\ref{table:modules}, `w/ PAM' and `w/ EAM' indicate that AAM is replaced with point attention module (PAM) and edge attention module (EAM), respectively. 
As revealed in the table, anisotropic attention leads to the best performance (12\% improvement in NME over the second leading method) among the three ones that attempt to make the model focus on salient regions.
As shown in Figure~\ref{figure:visualization}, the distribution of generated heatmap from AAM is aligned with the tangent direction rather than normal direction, which accounts for the effectiveness of AAM in our method. 

\begin{table}[t]
\begin{center}
\begin{tabular}{|l|c|c|c|}
\hline
Method & NME(\%) & FR$_{5\%}$ & AUC$_{5\%}$ \\
\hline
w/o AAM & 3.38 & 11.72 & 0.3653 \\
\hline
w/ PAM & 3.12 & 9.56 & 0.4095 \\
\hline
w/ EAM & 3.09 & 9.13 & 0.4124 \\
\hline
w/ AAM & \textbf{2.98} & \textbf{8.72} & \textbf{0.4318} \\
\hline
\end{tabular}
\end{center}
\caption{The contribution of AAM under different attention modules. 
ADL is not used in these settings.
}
\label{table:modules}
\vspace{-2.0em}
\end{table}

\vspace{5pt}
\noindent\textbf{Evaluation on different backbones in our method.}
To investigate the impact of backbones on face alignment performance, we trained models with two additional architectures.
All the settings are identical to ADNet except for the network architecture.
The results are tabulated in Table~\ref{table:backbones} where baseline denotes the model trained without ADL and AAM, in other words, baseline model feed $H_{landmarks}$ directly to soft-argmax, then apply $Smooth\ L_{1}$ (a.k.a $Smooth\ ADL_{1}$ with $\lambda$ = 1) loss to supervise landmark coordinates. 
For the ResNet50 backbone, we add three deconvolutional layers after the last layer to generate 64$\times$64 heatmap like our original model.
As shown in the table, all the models are significantly enhanced by equipping them with ADL and AAM, which demonstrates that ADL and AAM are generally adaptable regardless of network architecture.
In addition, ADNet with the default backbone (Stacked 4 HGNet) yields the best performance among three models.

\begin{table}[ht]
\begin{center}
\begin{tabular}{|l|c|c|}
\hline
Backbone & Baseline & ADNet \\
\hline
Stacked 4 HGNet \cite{newell2016stacked} & 3.38 & \textbf{2.93} \\
\hline
Single HGNet \cite{newell2016stacked} & 3.39 & \textbf{2.99} \\
\hline
ResNet50 \cite{he2016deep} & 3.43 & \textbf{3.11} \\
\hline
\end{tabular}
\end{center}
\caption{NME(\%) in different backbones on the 300W dataset.}
\label{table:backbones}
\end{table}

\noindent\textbf{Evaluation with and without facial attributes labels.}
It is shown that additional labeling information, such as face attributes, can help the model learn face alignment effectively. 
For example, PropNet~\cite{huang2020propagationnet} proposed a focal factor which dynamically adjusts the loss weight by utilizing class labels. 
We also test ADNet equipped with the focal factor, which is denoted as ADNet$^*$.
As shown in Table~\ref{table:WFLW} and \ref{table:attributes}, it is apparent that both PropNet~\cite{huang2020propagationnet} and ADNet$^*$ benefit from facial attribute labels.

\begin{table}[ht]
\begin{center}
\begin{tabular}{|l|c|c|c|}
\hline
Method & Tags & WFLW & 300W \\
\hline
PropNet & w/ & 4.05 & 2.93 \\
\hline
ADNet(Ours) & w/o & 4.14 & \textbf{2.93} \\
\hline
ADNet$^*$(Ours) & w/ & \textbf{3.98} & - \\
\hline
\end{tabular}
\end{center}
\caption{NME(\%) with and without attribute labels. The result of ADNet$^*$ on 300W is missing because the annotations for 300W was labelled by PropNet and the data is not published.}
\label{table:attributes}
\end{table}

\begin{figure}
\begin{center}
\includegraphics[width=3in]{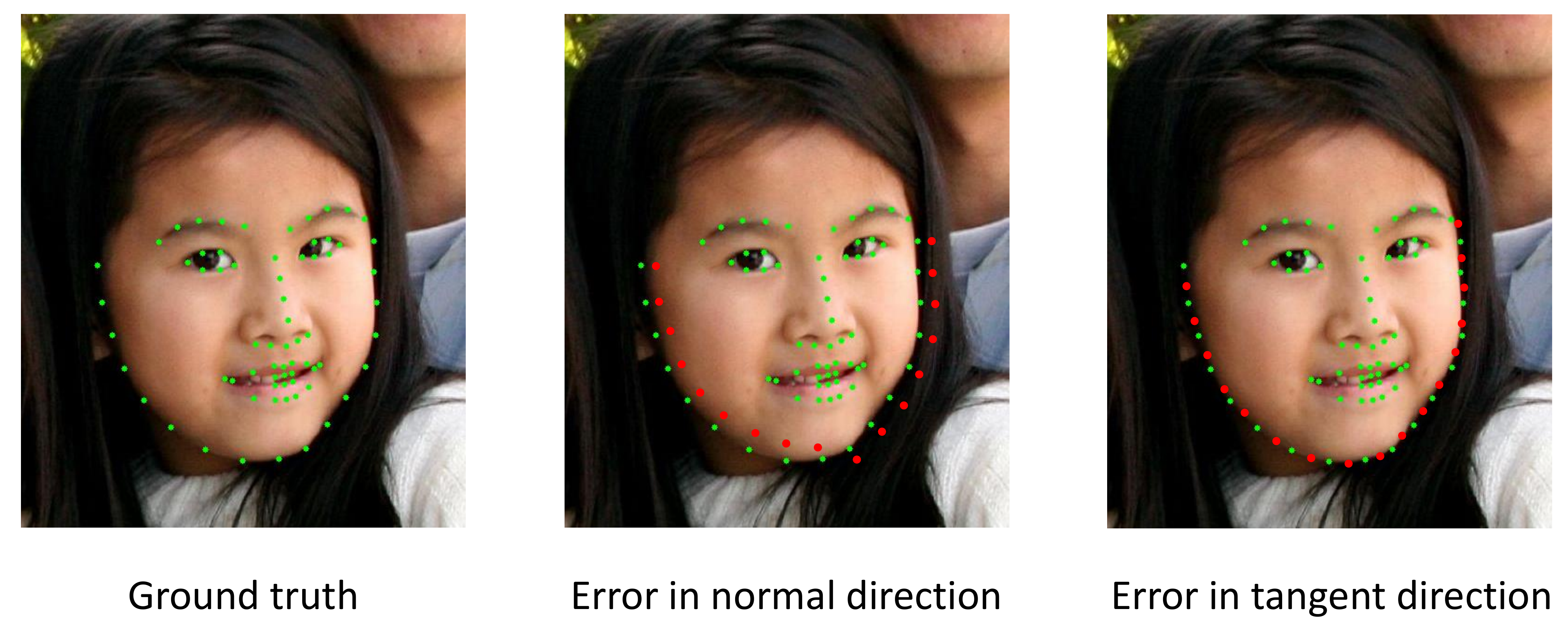}
\end{center}
\vspace{-1.0em}
   \caption{Visualization results on different error direction. Green points are ground truth, red points are mocked predictions. In the middle figure, all the mocked predictions have error in normal direction, and tangent direction in the right figure. Obviously, errors in tangent direction is more acceptable than normal direction which is aligned with labeling bias and human perception.}
\label{figure:N_T_better}
\vspace{-1.0em}
\end{figure}

\noindent\textbf{Evaluation on different error directions.}
Our fundamental assumption regarding the error-bias is that we should impose a strong constraint to the error along normal direction, while a relaxed constraint for the tangent-direction error.
Similar ideas were also discussed in previous studies, LAB \cite{wu2018look} and PropNet \cite{huang2020propagationnet}.
To further investigate the error direction, we report NME along normal and tangent directions independently as shown in Table~\ref{table:directions}
The baseline indicates the model without ADL and AAM. 
It can be observed that there is the significantly stronger NME bias along the tangent direction than tangent direction, which supports our assumption.
In addition, ADL and AAM appear to conduce to model accuracy enhancement, especially in normal direction, while allowing flexibility along the tangent direction to some degree, resulting in the increased bias rate of 48.05\%.
In other words, it is reasonable to impose a strong constraint to the error along the normal direction, which allows us to achieve better holistic performance in the end.
In addition, as demonstrated in Figure~\ref{figure:N_T_better}, errors along the tangent direction (the third image) are more acceptable to human perception than the same amount of the errors along the normal direction (the second image).

\begin{table}
\begin{center}
\begin{tabular}{|l|c|c|c|c|}
\hline
Method & \makecell{Normal \\ NME} & \makecell{Tangent \\ NME} & \makecell{Overall \\ NME} & Bias rate \\
\hline
Baseline & 1.91 & 2.55 & 3.38 & 33.51\% \\
\hline
ADNet & \textbf{1.54} & \textbf{2.28} & \textbf{2.93} & \textbf{48.05\%} \\
\hline
\end{tabular}
\end{center}
\caption{NME(\%) in normal and tangent directions of landmarks on the 300W dataset.}
\label{table:directions}
\end{table}

\begin{equation}
\textnormal{\emph{Bias Rate}} = \frac{\textnormal{\emph{NME}}_{tangent} - \textnormal{\emph{NME}}_{normal}}{\textnormal{\emph{NME}}_{normal}}
\label{equation:bias_rate}
\end{equation}

Where $\textnormal{\emph{NME}}_{tangent}$ and $\textnormal{\emph{NME}}_{normal}$ are respectively the NME in tangent and normal directions. 

\section{Conclusion}
\label{section:conclusion}

In this paper, we point out the significant bias of error distribution of each facial landmark, which is caused by ambiguity in the process of labeling.
Inspired by this knowledge, anisotropic direction loss is proposed to handle error in normal and tangent directions independently. Moreover, anisotropic attention module is put forward to efficiently combine point and edge heatmap information. 
By combining and integrating the two modules, the proposed model imposes more constraint on normal direction than on tangent direction for the learning of landmarks coordinates. And rigorous experiments show that the proposed method outperforms other state-of-the-art models on multiple datasets.
Finally, ablation studies verify the usefulness of the method comprehensively.
Additionally, we realized that the current metrics do not reflect the actual error bias issue, and result in the same error values for the $2$nd and $3$rd images in Figure~\ref{figure:N_T_better}.
In future work, the new metric considering human perception shown in Figure~\ref{figure:N_T_better} and the error-bias should be studied to provide more meaningful insights. 

{\small 
\bibliographystyle{ieee_fullname}
\bibliography{egbib}

\begin{thebibliography}{10}\itemsep=-1pt

\bibitem{bao2018towards}
Jianmin Bao, Dong Chen, Fang Wen, Houqiang Li, and Gang Hua.
\newblock Towards open-set identity preserving face synthesis.
\newblock In {\em Proceedings of the IEEE Conference on Computer Vision and
  Pattern Recognition}, pages 6713--6722, 2018.

\bibitem{bulat2016convolutional}
Adrian Bulat and Georgios Tzimiropoulos.
\newblock Convolutional aggregation of local evidence for large pose face
  alignment.
\newblock 2016.

\bibitem{bulat2017binarized}
Adrian Bulat and Georgios Tzimiropoulos.
\newblock Binarized convolutional landmark localizers for human pose estimation
  and face alignment with limited resources.
\newblock In {\em Proceedings of the IEEE International Conference on Computer
  Vision}, pages 3706--3714, 2017.

\bibitem{bulat2017far}
Adrian Bulat and Georgios Tzimiropoulos.
\newblock How far are we from solving the 2d \& 3d face alignment problem?(and
  a dataset of 230,000 3d facial landmarks).
\newblock In {\em Proceedings of the IEEE International Conference on Computer
  Vision}, pages 1021--1030, 2017.

\bibitem{buolamwini2018gender}
Joy Buolamwini and Timnit Gebru.
\newblock Gender shades: Intersectional accuracy disparities in commercial
  gender classification.
\newblock In {\em Conference on fairness, accountability and transparency},
  pages 77--91. PMLR, 2018.

\bibitem{burgos2013robust}
Xavier~P Burgos-Artizzu, Pietro Perona, and Piotr Doll{\'a}r.
\newblock Robust face landmark estimation under occlusion.
\newblock In {\em Proceedings of the IEEE international conference on computer
  vision}, pages 1513--1520, 2013.

\bibitem{cao2014face}
Xudong Cao, Yichen Wei, Fang Wen, and Jian Sun.
\newblock Face alignment by explicit shape regression.
\newblock {\em International Journal of Computer Vision}, 107(2):177--190,
  2014.

\bibitem{cootes2001active}
Timothy~F. Cootes, Gareth~J. Edwards, and Christopher~J. Taylor.
\newblock Active appearance models.
\newblock {\em IEEE Transactions on pattern analysis and machine intelligence},
  23(6):681--685, 2001.

\bibitem{cootes1992active}
Timothy~F Cootes and Christopher~J Taylor.
\newblock Active shape models—‘smart snakes’.
\newblock In {\em BMVC92}, pages 266--275. Springer, 1992.

\bibitem{cootes1995active}
Timothy~F Cootes, Christopher~J Taylor, David~H Cooper, and Jim Graham.
\newblock Active shape models-their training and application.
\newblock {\em Computer vision and image understanding}, 61(1):38--59, 1995.

\bibitem{dapogny2019decafa}
Arnaud Dapogny, Kevin Bailly, and Matthieu Cord.
\newblock Decafa: deep convolutional cascade for face alignment in the wild.
\newblock In {\em Proceedings of the IEEE/CVF International Conference on
  Computer Vision}, pages 6893--6901, 2019.

\bibitem{deng2019joint}
Jiankang Deng, George Trigeorgis, Yuxiang Zhou, and Stefanos Zafeiriou.
\newblock Joint multi-view face alignment in the wild.
\newblock {\em IEEE Transactions on Image Processing}, 28(7):3636--3648, 2019.

\bibitem{dong2018style}
Xuanyi Dong, Yan Yan, Wanli Ouyang, and Yi Yang.
\newblock Style aggregated network for facial landmark detection.
\newblock In {\em Proceedings of the IEEE Conference on Computer Vision and
  Pattern Recognition}, pages 379--388, 2018.

\bibitem{feng2015cascaded}
Zhen-Hua Feng, Guosheng Hu, Josef Kittler, William Christmas, and Xiao-Jun Wu.
\newblock Cascaded collaborative regression for robust facial landmark
  detection trained using a mixture of synthetic and real images with dynamic
  weighting.
\newblock {\em IEEE Transactions on Image Processing}, 24(11):3425--3440, 2015.

\bibitem{feng2018wing}
Zhen-Hua Feng, Josef Kittler, Muhammad Awais, Patrik Huber, and Xiao-Jun Wu.
\newblock Wing loss for robust facial landmark localisation with convolutional
  neural networks.
\newblock In {\em Proceedings of the IEEE Conference on Computer Vision and
  Pattern Recognition}, pages 2235--2245, 2018.

\bibitem{feng2017dynamic}
Zhen-Hua Feng, Josef Kittler, William Christmas, Patrik Huber, and Xiao-Jun Wu.
\newblock Dynamic attention-controlled cascaded shape regression exploiting
  training data augmentation and fuzzy-set sample weighting.
\newblock In {\em Proceedings of the IEEE conference on computer vision and
  pattern recognition}, pages 2481--2490, 2017.

\bibitem{gu2019mask}
Shuyang Gu, Jianmin Bao, Hao Yang, Dong Chen, Fang Wen, and Lu Yuan.
\newblock Mask-guided portrait editing with conditional gans.
\newblock In {\em Proceedings of the IEEE/CVF Conference on Computer Vision and
  Pattern Recognition}, pages 3436--3445, 2019.

\bibitem{he2016deep}
Kaiming He, Xiangyu Zhang, Shaoqing Ren, and Jian Sun.
\newblock Deep residual learning for image recognition.
\newblock In {\em Proceedings of the IEEE conference on computer vision and
  pattern recognition}, pages 770--778, 2016.

\bibitem{honari2018improving}
Sina Honari, Pavlo Molchanov, Stephen Tyree, Pascal Vincent, Christopher Pal,
  and Jan Kautz.
\newblock Improving landmark localization with semi-supervised learning.
\newblock In {\em Proceedings of the IEEE Conference on Computer Vision and
  Pattern Recognition}, pages 1546--1555, 2018.

\bibitem{huang2017densely}
Gao Huang, Zhuang Liu, Laurens Van Der~Maaten, and Kilian~Q Weinberger.
\newblock Densely connected convolutional networks.
\newblock In {\em Proceedings of the IEEE conference on computer vision and
  pattern recognition}, pages 4700--4708, 2017.

\bibitem{huang2020ace}
Jihua Huang and Amir Tamrakar.
\newblock Ace-net: Fine-level face alignment through anchors and contours
  estimation.
\newblock {\em arXiv preprint arXiv:2012.01461}, 2020.

\bibitem{huang2020propagationnet}
Xiehe Huang, Weihong Deng, Haifeng Shen, Xiubao Zhang, and Jieping Ye.
\newblock Propagationnet: Propagate points to curve to learn structure
  information.
\newblock In {\em Proceedings of the IEEE/CVF Conference on Computer Vision and
  Pattern Recognition}, pages 7265--7274, 2020.

\bibitem{jiang2005efficient}
Dalong Jiang, Yuxiao Hu, Shuicheng Yan, Lei Zhang, Hongjiang Zhang, and Wen
  Gao.
\newblock Efficient 3d reconstruction for face recognition.
\newblock {\em Pattern Recognition}, 38(6):787--798, 2005.

\bibitem{kahraman2007active}
Fatih Kahraman, Muhittin Gokmen, Sune Darkner, and Rasmus Larsen.
\newblock An active illumination and appearance (aia) model for face alignment.
\newblock In {\em 2007 IEEE Conference on Computer Vision and Pattern
  Recognition}, pages 1--7. IEEE, 2007.

\bibitem{kumar2018disentangling}
Amit Kumar and Rama Chellappa.
\newblock Disentangling 3d pose in a dendritic cnn for unconstrained 2d face
  alignment.
\newblock In {\em Proceedings of the IEEE Conference on Computer Vision and
  Pattern Recognition}, pages 430--439, 2018.

\bibitem{kumar2020luvli}
Abhinav Kumar, Tim~K Marks, Wenxuan Mou, Ye Wang, Michael Jones, Anoop Cherian,
  Toshiaki Koike-Akino, Xiaoming Liu, and Chen Feng.
\newblock Luvli face alignment: Estimating landmarks' location, uncertainty,
  and visibility likelihood.
\newblock In {\em Proceedings of the IEEE/CVF Conference on Computer Vision and
  Pattern Recognition}, pages 8236--8246, 2020.

\bibitem{liu2018intriguing}
Rosanne Liu, Joel Lehman, Piero Molino, Felipe~Petroski Such, Eric Frank, Alex
  Sergeev, and Jason Yosinski.
\newblock An intriguing failing of convolutional neural networks and the
  coordconv solution.
\newblock {\em arXiv preprint arXiv:1807.03247}, 2018.

\bibitem{lv2017deep}
Jiangjing Lv, Xiaohu Shao, Junliang Xing, Cheng Cheng, and Xi Zhou.
\newblock A deep regression architecture with two-stage re-initialization for
  high performance facial landmark detection.
\newblock In {\em Proceedings of the IEEE conference on computer vision and
  pattern recognition}, pages 3317--3326, 2017.

\bibitem{masi2018deep}
Iacopo Masi, Yue Wu, Tal Hassner, and Prem Natarajan.
\newblock Deep face recognition: A survey.
\newblock In {\em 2018 31st SIBGRAPI conference on graphics, patterns and
  images (SIBGRAPI)}, pages 471--478. IEEE, 2018.

\bibitem{matthews2004active}
Iain Matthews and Simon Baker.
\newblock Active appearance models revisited.
\newblock {\em International journal of computer vision}, 60(2):135--164, 2004.

\bibitem{miao2018direct}
Xin Miao, Xiantong Zhen, Xianglong Liu, Cheng Deng, Vassilis Athitsos, and Heng
  Huang.
\newblock Direct shape regression networks for end-to-end face alignment.
\newblock In {\em Proceedings of the IEEE Conference on Computer Vision and
  Pattern Recognition}, pages 5040--5049, 2018.

\bibitem{milborrow2008locating}
Stephen Milborrow and Fred Nicolls.
\newblock Locating facial features with an extended active shape model.
\newblock In {\em European conference on computer vision}, pages 504--513.
  Springer, 2008.

\bibitem{newell2016stacked}
Alejandro Newell, Kaiyu Yang, and Jia Deng.
\newblock Stacked hourglass networks for human pose estimation.
\newblock In {\em European conference on computer vision}, pages 483--499.
  Springer, 2016.

\bibitem{ren2014face}
Shaoqing Ren, Xudong Cao, Yichen Wei, and Jian Sun.
\newblock Face alignment at 3000 fps via regressing local binary features.
\newblock In {\em Proceedings of the IEEE Conference on Computer Vision and
  Pattern Recognition}, pages 1685--1692, 2014.

\bibitem{ronneberger2015u}
Olaf Ronneberger, Philipp Fischer, and Thomas Brox.
\newblock U-net: Convolutional networks for biomedical image segmentation.
\newblock In {\em International Conference on Medical image computing and
  computer-assisted intervention}, pages 234--241. Springer, 2015.

\bibitem{sagonas2016300}
Christos Sagonas, Epameinondas Antonakos, Georgios Tzimiropoulos, Stefanos
  Zafeiriou, and Maja Pantic.
\newblock 300 faces in-the-wild challenge: Database and results.
\newblock {\em Image and vision computing}, 47:3--18, 2016.

\bibitem{sagonas2013300}
Christos Sagonas, Georgios Tzimiropoulos, Stefanos Zafeiriou, and Maja Pantic.
\newblock 300 faces in-the-wild challenge: The first facial landmark
  localization challenge.
\newblock In {\em Proceedings of the IEEE International Conference on Computer
  Vision Workshops}, pages 397--403, 2013.

\bibitem{saragih2007nonlinear}
Jason Saragih and Roland Goecke.
\newblock A nonlinear discriminative approach to aam fitting.
\newblock In {\em 2007 IEEE 11th International Conference on Computer Vision},
  pages 1--8. IEEE, 2007.

\bibitem{sauer2011accurate}
Patrick Sauer, Timothy~F Cootes, and Christopher~J Taylor.
\newblock Accurate regression procedures for active appearance models.
\newblock In {\em BMVC}, pages 1--11, 2011.

\bibitem{sun2013deep}
Yi Sun, Xiaogang Wang, and Xiaoou Tang.
\newblock Deep convolutional network cascade for facial point detection.
\newblock In {\em Proceedings of the IEEE conference on computer vision and
  pattern recognition}, pages 3476--3483, 2013.

\bibitem{tang2018quantized}
Zhiqiang Tang, Xi Peng, Shijie Geng, Lingfei Wu, Shaoting Zhang, and Dimitris
  Metaxas.
\newblock Quantized densely connected u-nets for efficient landmark
  localization.
\newblock In {\em Proceedings of the European Conference on Computer Vision
  (ECCV)}, pages 339--354, 2018.

\bibitem{toshev2014deeppose}
Alexander Toshev and Christian Szegedy.
\newblock Deeppose: Human pose estimation via deep neural networks.
\newblock In {\em Proceedings of the IEEE conference on computer vision and
  pattern recognition}, pages 1653--1660, 2014.

\bibitem{trigeorgis2016mnemonic}
George Trigeorgis, Patrick Snape, Mihalis~A Nicolaou, Epameinondas Antonakos,
  and Stefanos Zafeiriou.
\newblock Mnemonic descent method: A recurrent process applied for end-to-end
  face alignment.
\newblock In {\em Proceedings of the IEEE Conference on Computer Vision and
  Pattern Recognition}, pages 4177--4187, 2016.

\bibitem{valle2018deeply}
Roberto Valle, Jose~M Buenaposada, Antonio Valdes, and Luis Baumela.
\newblock A deeply-initialized coarse-to-fine ensemble of regression trees for
  face alignment.
\newblock In {\em Proceedings of the European Conference on Computer Vision
  (ECCV)}, pages 585--601, 2018.

\bibitem{wang2019adaptive}
Xinyao Wang, Liefeng Bo, and Li Fuxin.
\newblock Adaptive wing loss for robust face alignment via heatmap regression.
\newblock In {\em Proceedings of the IEEE/CVF International Conference on
  Computer Vision}, pages 6971--6981, 2019.

\bibitem{wei2016convolutional}
Shih-En Wei, Varun Ramakrishna, Takeo Kanade, and Yaser Sheikh.
\newblock Convolutional pose machines.
\newblock In {\em Proceedings of the IEEE conference on Computer Vision and
  Pattern Recognition}, pages 4724--4732, 2016.

\bibitem{wu2018look}
Wayne Wu, Chen Qian, Shuo Yang, Quan Wang, Yici Cai, and Qiang Zhou.
\newblock Look at boundary: A boundary-aware face alignment algorithm.
\newblock In {\em Proceedings of the IEEE conference on computer vision and
  pattern recognition}, pages 2129--2138, 2018.

\bibitem{wu2017leveraging}
Wenyan Wu and Shuo Yang.
\newblock Leveraging intra and inter-dataset variations for robust face
  alignment.
\newblock In {\em Proceedings of the IEEE conference on computer vision and
  pattern recognition workshops}, pages 150--159, 2017.

\bibitem{wu2015robust}
Yue Wu and Qiang Ji.
\newblock Robust facial landmark detection under significant head poses and
  occlusion.
\newblock In {\em Proceedings of the IEEE International Conference on Computer
  Vision}, pages 3658--3666, 2015.

\bibitem{xiao2016robust}
Shengtao Xiao, Jiashi Feng, Junliang Xing, Hanjiang Lai, Shuicheng Yan, and
  Ashraf Kassim.
\newblock Robust facial landmark detection via recurrent attentive-refinement
  networks.
\newblock In {\em European conference on computer vision}, pages 57--72.
  Springer, 2016.

\bibitem{xiong2013supervised}
Xuehan Xiong and Fernando De~la Torre.
\newblock Supervised descent method and its applications to face alignment.
\newblock In {\em Proceedings of the IEEE conference on computer vision and
  pattern recognition}, pages 532--539, 2013.

\bibitem{zhang2014coarse}
Jie Zhang, Shiguang Shan, Meina Kan, and Xilin Chen.
\newblock Coarse-to-fine auto-encoder networks (cfan) for real-time face
  alignment.
\newblock In {\em European conference on computer vision}, pages 1--16.
  Springer, 2014.

\bibitem{zhang2019making}
Richard Zhang.
\newblock Making convolutional networks shift-invariant again.
\newblock In {\em International Conference on Machine Learning}, pages
  7324--7334. PMLR, 2019.

\bibitem{zhang2014facial}
Zhanpeng Zhang, Ping Luo, Chen~Change Loy, and Xiaoou Tang.
\newblock Facial landmark detection by deep multi-task learning.
\newblock In {\em European conference on computer vision}, pages 94--108.
  Springer, 2014.

\bibitem{zhang2015learning}
Zhanpeng Zhang, Ping Luo, Chen~Change Loy, and Xiaoou Tang.
\newblock Learning deep representation for face alignment with auxiliary
  attributes.
\newblock {\em IEEE transactions on pattern analysis and machine intelligence},
  38(5):918--930, 2015.

\bibitem{zhou2013extensive}
Erjin Zhou, Haoqiang Fan, Zhimin Cao, Yuning Jiang, and Qi Yin.
\newblock Extensive facial landmark localization with coarse-to-fine
  convolutional network cascade.
\newblock In {\em Proceedings of the IEEE international conference on computer
  vision workshops}, pages 386--391, 2013.

\bibitem{zhu2015face}
Shizhan Zhu, Cheng Li, Chen Change~Loy, and Xiaoou Tang.
\newblock Face alignment by coarse-to-fine shape searching.
\newblock In {\em Proceedings of the IEEE conference on computer vision and
  pattern recognition}, pages 4998--5006, 2015.

\end{thebibliography}
}

\clearpage
\appendix


\section{Appendix}
\label{section:appendix}


\subsection{Model Architecture}
Tables~\ref{table:architecture_ADNet},~\ref{table:architecture_head_branches}~and~\ref{table:architecture_residual_block} fully demonstrate the architecture of the proposed ADNet. For detailed introduction of our experimental setting, please refer to Section~4 of our manuscript. In the table, $P*$, $H*_{point}$ and $H*_{edge}$ denote the inputs of \emph{smooth ADL1} loss, \emph{AWing} loss and \emph{AWing} loss, respectively.
$N_{point}$ and $N_{edge}$ indicate the number of points and edges, which varies according to each dataset.
The loss weights of Hour Glass (HG) for stacked 4 HGs are respectively 1/8, 1/4, 1/2, and 1.
The fourth head branch outputs $P_3$ is the final predicted coordinate of each landmark, which is derived from the soft argmax operation.

In Table~\ref{table:architecture_head_branches}, the goal of $E2P\ \textnormal{\emph{Transform}}$ is to convert $\hat{H}_{edge}$ ($N_{edge}$ channels) into $H_{edge}$ ($N_{point}$ channels) by considering the adjacency relationship as
\begin{equation}
\small
E2P\ \textnormal{\emph{Transform}}(\hat{H}_{edge}(x, y)) = \textnormal{\emph{Mat}}_{E2P} \cdot \hat{H}_{edge}(x, y)
\label{equation:E2P_transform}
\end{equation}
where $\hat{H}_{edge}(x, y)$ is a column vector at the position of $(x, y)$, and $\textnormal{\emph{Mat}}_{E2P}$ is a $N_{point}{\times}N_{edge}$ binary matrix describing the adjacency relationship between each point and each edge.
More specifically, if the $i$th point is connected to the $j$th edge, $\textnormal{\emph{Mat}}_{E2P}(i, j)=1$, otherwise, $\textnormal{\emph{Mat}}_{E2P}(i, j)=0$.
Note that $\textnormal{\emph{Mat}}_{E2P}$ is a constant variable, and is derived based on the landmark definition of each dataset, respectively.

\begin{table*}[htbp]
\centering
\small
\begin{subtable}
\centering
\begin{tabular}{|p{0.95in}|p{0.85in}|p{1.5in}|p{1.0in}|p{0.5in}|p{0.5in}|p{0.5in}|}
\hline
Layer & Input of layer & Output of layer & Output Channels & Kernel Size & Stride & Padding \\
\hline
Input & image & - & - & - & - & - \\
\hline
Coord Conv~\cite{liu2018intriguing} & image & x0 & 64 & 7 & 2 & 3 \\
BN-ReLu & x0 & x1 & 64 & - & - & - \\
Residual Block~\cite{he2016deep} & x1 & x2 & 128 & - & - & - \\
Max Pool & x2 & x3 & 128 & 2 & 2 & 0 \\
Blur Pool~\cite{zhang2019making} & x3 & x4 & 128 & 3 & 2 & 0 \\
Residual Block & x4 & x5 & 128 & - & - & - \\
Residual Block & x5 & x6 & 256 & - & - & - \\
\hline
Head Branch & x6 & ($P0$, x7, $H0_{point}$, $H0_{edge}$) & - & - & - & - \\
Head Branch & x7 & ($P1$, x8, $H1_{point}$, $H1_{edge}$) & - & - & - & - \\
Head Branch & x8 & ($P2$, x9, $H2_{point}$, $H2_{edge}$) & - & - & - & - \\
Head Branch & x9 & ($P3$, x10, $H3_{point}$, $H3_{edge}$) & - & - & - & - \\
\hline
Output & - & ($P*$, $H*_{point}$, $H*_{edge}$) & - & - & - & - \\
\hline
\end{tabular}
\caption{The architecture of ADNet. x[*] and $H$[*] indicate intermediate feature maps, and BN indicates batch normalization. The detailed structure of ``Head Branch'' and ``Residual Block'' are shown in Tables~\ref{table:architecture_head_branches}~and~\ref{table:architecture_residual_block}.}
\label{table:architecture_ADNet}
\end{subtable}

\hfill
\bigskip
\break

\begin{subtable}
\centering
\begin{tabular}{|p{0.95in}|p{0.85in}|p{1.5in}|p{1.0in}|p{0.5in}|p{0.5in}|p{0.5in}|}
\hline
Layer & Input of layer & Output of layer & Output Channels & Kernel Size & Stride & Padding \\
\hline
Input & y0 & - & - & - & - & - \\
\hline
Hour Glass~\cite{newell2016stacked} & y0 & y1 & 256 & - & - & - \\
Conv-BN-ReLu & y1 & y2 & 256 & 1 & 1 & 0 \\
Residual Block & y2 & y3 & 256 & - & - & - \\
\hline
Conv-Sigmoid & y3 & $H_{point}$ & $N_{point}$ & 1 & 1 & 0 \\
Conv-Sigmoid & y3 & $\hat{H}_{edge}$ & $N_{edge}$ & 1 & 1 & 0 \\
E2P Transform & $\hat{H}_{edge}$ & $H_{edge}$ & $N_{point}$ & - & - & - \\
Elementwise dot & ($H_{point}$, $H_{edge}$) & $H_{point-edge}$ & $N_{point}$ & - & - & - \\
\hline
Conv-ReLu & y3 & $H_{landmarks}$ & $N_{point}$ & 1 & 1 & 0 \\
Elementwise dot & ($H_{landmarks}$, $H_{point-edge}$) & $AH_{landmarks}$ & $N_{point}$ & - & - & - \\
Soft Argmax & $AH_{landmarks}$ & $P$ & $N_{point}$ & - & - & - \\
\hline
Conv & $H_{landmarks}$ & y4 & 256 & 1 & 1 & 0 \\
Conv & $H_{point}$ & y5 & 256 & 1 & 1 & 0 \\
Conv & $H_{edge}$ & y6 & 256 & 1 & 1 & 0 \\
Elementwise sum & (y3, y4, y5, y6) & y7 & 256 & - & - & - \\
\hline
Output & - & ($P$, y7, $H_{point}$, $H_{edge}$) & - & - & - & - \\
\hline
\end{tabular}
\caption{The architecture of head branch.}
\label{table:architecture_head_branches}
\end{subtable}

\hfill
\bigskip
\break

\begin{subtable}
\centering
\begin{tabular}{|p{0.95in}|p{0.85in}|p{1.5in}|p{1.0in}|p{0.5in}|p{0.5in}|p{0.5in}|}
\hline
Layer & Input of layer & Output of layer & Output Channels & Kernel Size & Stride & Padding \\
\hline
Input & z0 & - & - & - & - & - \\
\hline
BN-ReLu-Conv & z0 & z1 & output channels / 2 & 1 & 1 & 0 \\
BN-ReLu-Conv & z1 & z2 & output channels / 2 & 3 & 1 & 1 \\
BN-ReLu-Conv & z2 & z3 & output channels & 1 & 1 & 0 \\
\hline
Skip & z0 & z4 & output channels & 1 & 1 & 0 \\
Elementwise sum & (z3, z4) & z5 & output channels & 1 & 1 & 0 \\
\hline
Output & - & z5 & - & - & - & - \\
\hline
\end{tabular}
\caption{The architecture of residual block. ``output channels'' denotes the channel size of the residual block's output.}
\label{table:architecture_residual_block}
\end{subtable}

\vspace{5mm}

\end{table*}

\subsection{Edge Definition}
\begin{figure}[htbp]
\begin{center}
\includegraphics[width=3.1in]{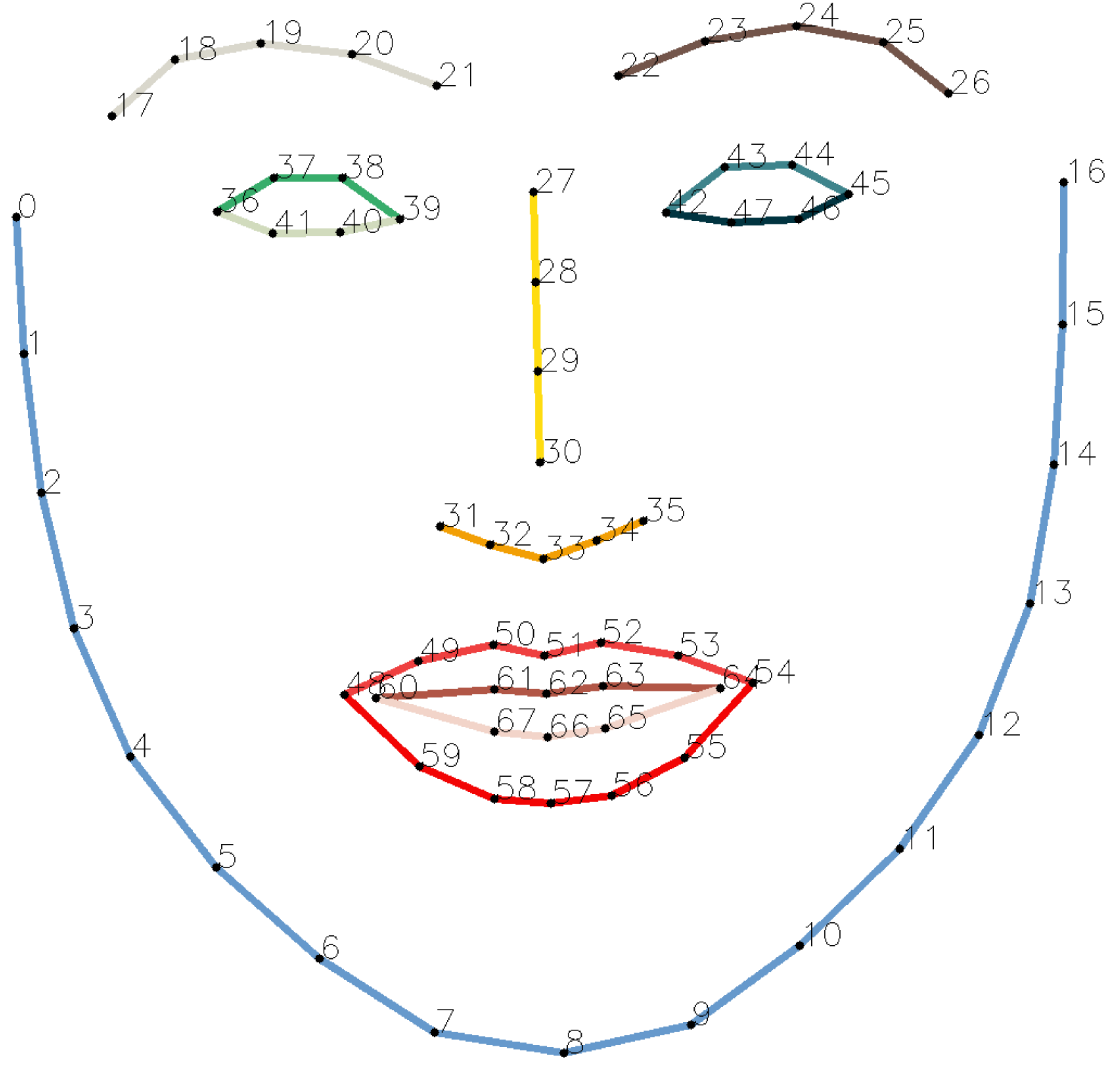}
\end{center}
\vspace{-5mm}
\caption{Visualized example of edges in 300W. Each colored line corresponds to each edge defined in Table~\ref{table:edge_definition}.}
\label{figure:edge_definition}
\end{figure}

We categorize the landmarks into two groups: \emph{edge landmarks} and \emph{point landmarks}. If the landmarks locate on edges, they belong to the former group, conversely, landmarks not on edges belong to the latter group. For several well-known face alignment datasets such as COFW, 300W, and WFLW, most of the landmarks belong to edge landmarks. 
We show our definition of edges in 300W dataset in Table~\ref{table:edge_definition} and Figure~\ref{figure:edge_definition}.

\begin{table}[htbp]
\begin{center}
\small
\begin{tabular}{|l|c|c|}
\hline
Components & Edge Names & Vertex Indices \\
\hline
\multirow{1}*{Contour} & \textbf{\textcolor[rgb]{0.4,0.6,0.8}{Face Contour}} & 0-16 \\
\hline
\multirow{2}*{Eyebrow} & \textbf{\textcolor[rgb]{0.85,0.843,0.8}{Right Eyebrow}} & 17-21 \\
 & \textbf{\textcolor[rgb]{0.6,0.33,0.29}{Left Eyebrow}} & 22-26 \\
\hline
\multirow{2}*{Nose} & \textbf{\textcolor[rgb]{0.99,0.863,0.067}{Nose Middle Line}} & 27-30 \\
 & \textbf{\textcolor[rgb]{0.95,0.623,0.0197}{Nose Bottom Line}} & 31-35 \\
\hline
\multirow{4}*{Eye} & \textbf{\textcolor[rgb]{0.2156,0.6784,0.4196}{Right Eye Superior Margin}} & 36-39 \\
 & \textbf{\textcolor[rgb]{0.8196,0.8589,0.7412}{Right Eye Inferior Margin}} & 39-41, 36 \\
 & \textbf{\textcolor[rgb]{0.243,0.5176,0.549}{Left Eye Superior Margin}} & 42-45 \\
 & \textbf{\textcolor[rgb]{0.004,0.204,0.25}{Left Eye Inferior Margin}} & 45-47, 42 \\
\hline
\multirow{4}*{Mouth} & \textbf{\textcolor[rgb]{0.937,0.25,0.25}{Outer Lip Superior Margin}} & 48-54 \\
 & \textbf{\textcolor[rgb]{0.949,0.0196,0.0196}{Outer Lip Inferior Margin}} & 54-59, 48 \\
 & \textbf{\textcolor[rgb]{0.694,0.3373,0.2745}{Inner Lip Superior Margin}} & 60-64 \\
 & \textbf{\textcolor[rgb]{0.949,0.831,0.788}{Inner Lip Inferior Margin}} & 64-67, 60 \\
\hline
Whole face & - & 0-67 \\
\hline
\end{tabular}
\end{center}
\vspace{-2mm}
\caption{Definition of edges in 300W. The visualized example of each edge is shown in Figure~\ref{figure:edge_definition} with the same color.}
\label{table:edge_definition}
\end{table}

\subsection{Additional Experiments and Results}

\subsubsection{Comparison of Inference Time}
To show the computational complexity of ADL and AAM, we compare the inference time of the baseline model and ADNet. Note that the baseline model is almost identical to ADNet except that AAM and ADL are removed from the baseline.
To estimate the time, we repeated the experiment 10 times on the 300W fullset and averaged the measured times. 
We used one NVIDIA v100 GPU with a batch size of 1.
As tabulated in Table~\ref{table:exp_time_cost}, ADNet takes only 6\% longer time than the baseline method, which indicates the high efficiency of ADL and AAM.
Moreover, ADL and AAM take small FLOPs and require a small number of parameters as shown in the table.

\begin{table}[htbp]
\centering
\begin{tabular}{|c|c|c|c|}
\hline
Methods & Inference Time & FLOPs & Params \\
\hline
Baseline & 89.49 ms/face & 16.46G & 13.23M \\
ADNet & 95.29 ms/face & 17.04G & 13.37M \\
\hline
\end{tabular}
\vspace{1mm}
\caption{The comparison of inference time, FLOPs and the number of parameters on the 300W fullset.}
\label{table:exp_time_cost}
\end{table}

\subsubsection{Evaluation of Individual Edges on 300W}
Apart from evaluating the whole face on the test dataset, we also provide the NME of each edge in the 300W fullset dataset to fully demonstrate the effectiveness of the proposed method. 
The detailed results are shown in Table~\ref{table:Edges_test_300W}. The bias rate is defined as
\begin{equation}
\textnormal{\emph{Bias Rate}} = \frac{\textnormal{\emph{NME}}_{tangent} - \textnormal{\emph{NME}}_{normal}}{\textnormal{\emph{NME}}_{normal}}
\label{equation:bias_rate}
\end{equation}
where $\textnormal{\emph{NME}}_{tangent}$ and $\textnormal{\emph{NME}}_{normal}$ are respectively the NME in tangent and normal directions. 
For both normal NME and tangent NME, ADNet outperforms the baseline method for every edge. In addition, ADNet has always larger bias rate than the baseline, which means that ADNet is leveraging the bias towards normal direction.

\subsubsection{Exploration of $\lambda$ Settings}

We investigate three $\lambda$ settings in Table~\ref{table:diff_lambda}:
\textbf{i)} All landmarks have the same value $\lambda_i=2$: (c)(f). 
Other $\lambda_i$ can be found in Table~\ref{table:diff_lambda}.
\textbf{ii)} $\lambda_i=4$ for the outer face contour (denoted by $\mathcal{O}$ in Table~\ref{table:diff_lambda}), and $\lambda_i=2$ for the rest: (d)(g).
\textbf{iii)} Independent $\lambda_i$ for each landmark: (e)(h).
Each was computed by $\lambda_i=a_i/{b_i}$, where $a_i$ and $b_i$ are long and short radius of each fitted ellipse by error distribution in Figure~\ref{figure:error_distribution}(a).

It can be observed that: 
\textbf{i)} though a more flexible $\lambda_i$ leads to better performance, the improvement is marginal; 
\textbf{ii)} the significant improvement comes from AAM rather than ADL.

\begin{table}[ht]
\footnotesize
\begin{center}
\begin{tabular}{l|c|l|l}
\hline
ID & Components & $\lambda_i$ & NME (\%) \\
\hline
(a) & \emph{Baseline} & - & 3.38 \\
(b) & AAM only & - & 2.98 \\
\hline
(c) & ADL only & $\lambda_i=2$ & 3.231951 \\
(d) & ADL only & $\lambda_{i \in \mathcal{O}}=4$, $\lambda_{i \not\in \mathcal{O}}=2$ & 3.229207 \\
(e) & ADL only & $\lambda_i=a_i/{b_i}$ & 3.219207 \\
\hline
(f) & AAM + ADL & $\lambda_i=2$ & 2.934116 \\
(g) & AAM + ADL & $\lambda_{i \in \mathcal{O}}=4$, $\lambda_{i \not\in \mathcal{O}}=2$ & 2.934933 \\
(h) & AAM + ADL & $\lambda_i=a_i/{b_i}$ & 2.930612 \\
\hline
\end{tabular}
\vspace{-2mm}
\end{center}
\caption{Evaluating different $\lambda$ strategies on 300W in terms of interocular NME. The \emph{Baseline} in (a) removes both AAM and ADL.}
\label{table:diff_lambda}
\vspace{-5mm}
\end{table}

\subsubsection{Demonstration of Error Distribution on 300W}
To demonstrate the error-bias in error distribution with real-world data, in Figure~\ref{figure:error_distribution}, we provide the empirical error distribution of chin point obtained by using an off-the-shelf face alignment algorithm on the 300W dataset trained by baseline method.
It is obvious that the error distribution along tangent direction (tangent distribution in figure) is broader than that along the normal direction (normal distribution in figure), which is consistent with our assumption, error-bias towards normal direction.

\begin{figure}[htbp]
\begin{center}
\includegraphics[width=3.2in]{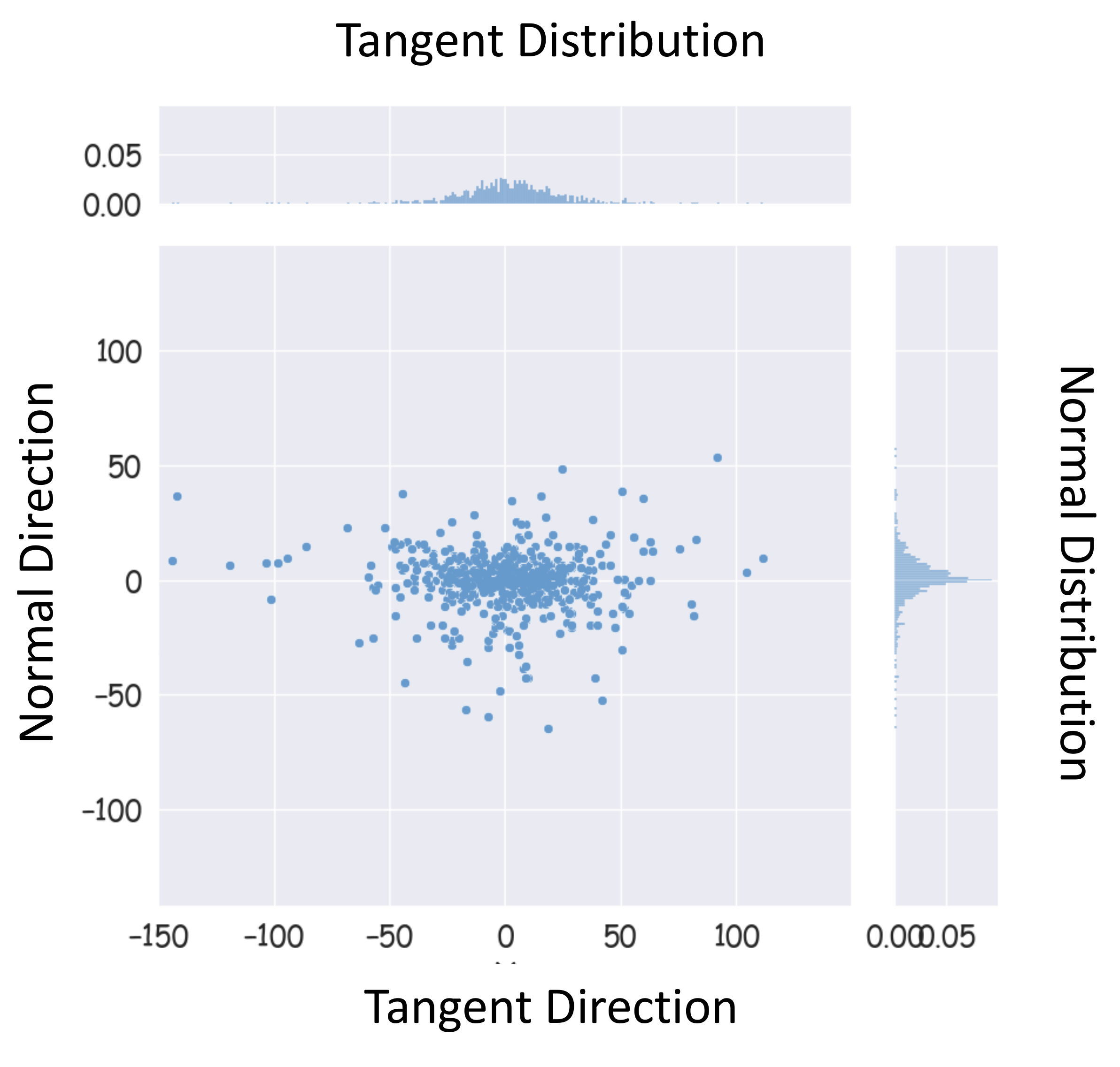}
\end{center}
\vspace{-6mm}
\caption{Error distribution of chin landmark (the $8th$ point in Figures~\ref{figure:edge_definition}) on the 300W fullset dataset obtained by off-the-shelf face alignment model. Each sub-figure (up/right) shows the projected error distribution along (tangent/normal) direction.}
\label{figure:error_distribution}
\vspace{-6mm}
\end{figure}

\subsubsection{Visualized Examples of ADNet}
\begin{figure*}[htbp]
\begin{center}
\includegraphics[width=6.8in]{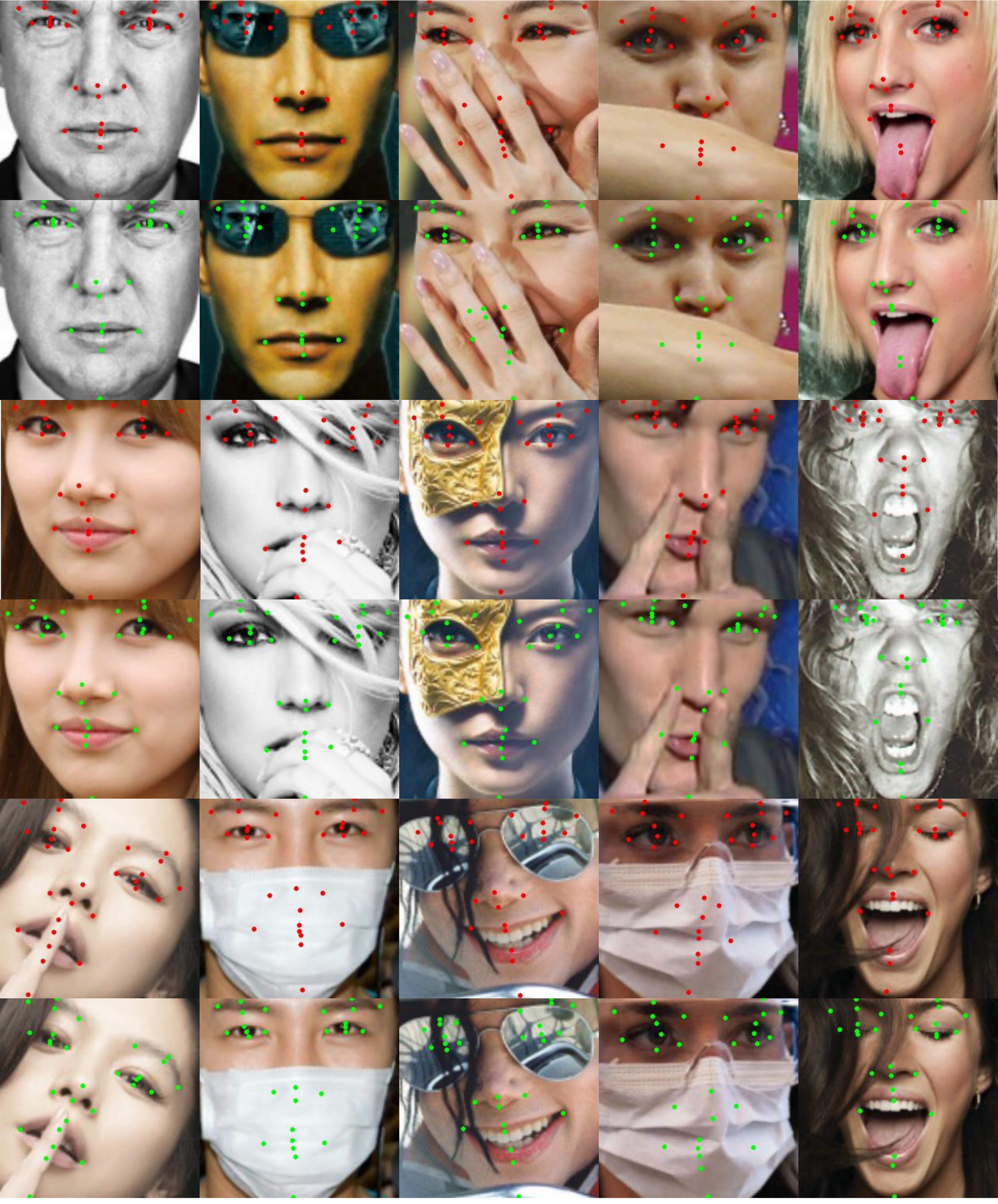}
\end{center}
\caption{Visualized examples in COFW test dataset. (\textcolor[rgb]{1.00,0.00,0.00}{Red} denotes predicted values by ADNet model and \textcolor[rgb]{0.00,1.00,0.00}{Green} denotes ground truth.)}
\label{figure:visualization_result_COFW}
\end{figure*}

\begin{figure*}[htbp]
\begin{center}
\includegraphics[width=6.8in]{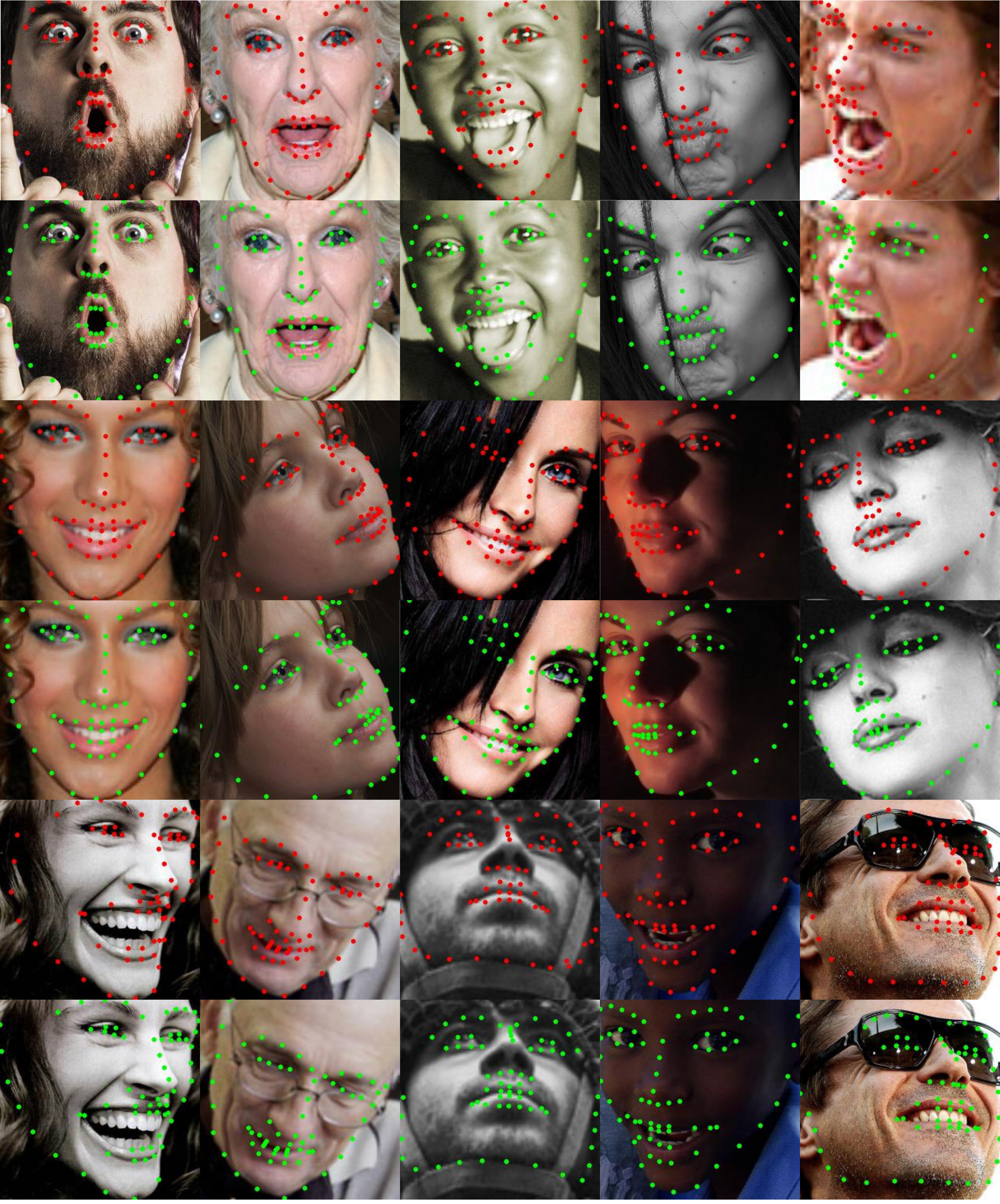}
\end{center}
\caption{Visualized examples in the 300W test dataset. (\textcolor[rgb]{1.00,0.00,0.00}{Red} denotes predicted values by ADNet model and \textcolor[rgb]{0.00,1.00,0.00}{Green} denotes ground truth.)}
\label{figure:visualization_result_300W}
\end{figure*}

\begin{figure*}[htbp]
\begin{center}
\includegraphics[width=6.8in]{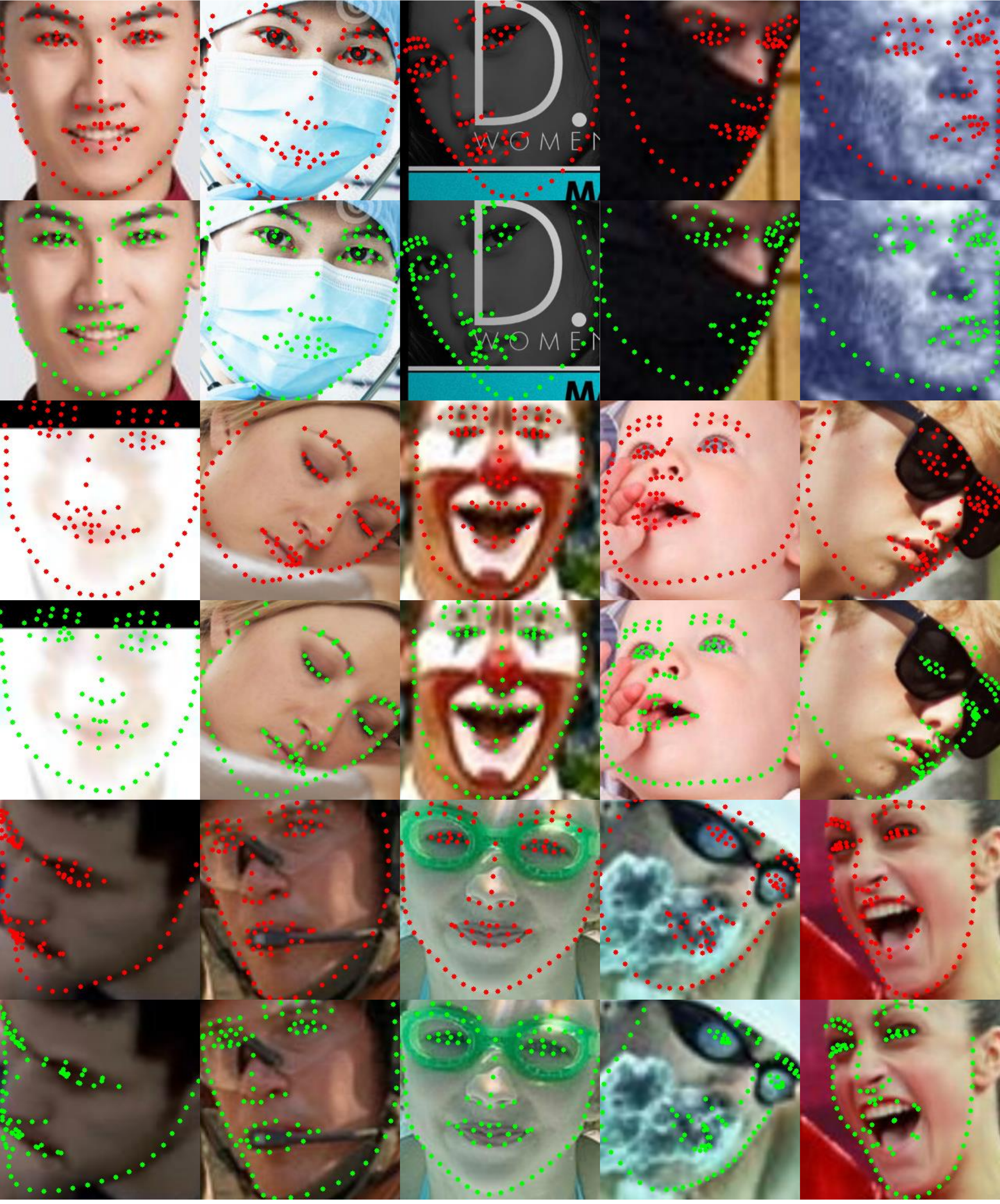}
\end{center}
\caption{Visualized examples in the WFLW test dataset. (\textcolor[rgb]{1.00,0.00,0.00}{Red} denotes predicted values by ADNet model and \textcolor[rgb]{0.00,1.00,0.00}{Green} denotes ground truth.)}
\label{figure:visualization_result_WFLW}
\end{figure*}

To verify the robustness of ADNet, we additionally show the landmark inference on the extended test data in Figure~\ref{figure:visualization_result_COFW},~\ref{figure:visualization_result_300W}~and~\ref{figure:visualization_result_WFLW}. 
For each image, the first row (red landmarks) is the inference result by ADNet and the second row (green landmarks) is the corresponding ground-truth provided by the dataset.
As can be seen, our method yields stable and reasonable prediction of landmarks even for difficult cases such as extreme occlusion, large pose, extreme expression, blur and bad illumination.

\subsection{Extra Description of AAM}

\begin{figure}[htbp]
\begin{center}
\includegraphics[width=3.2in]{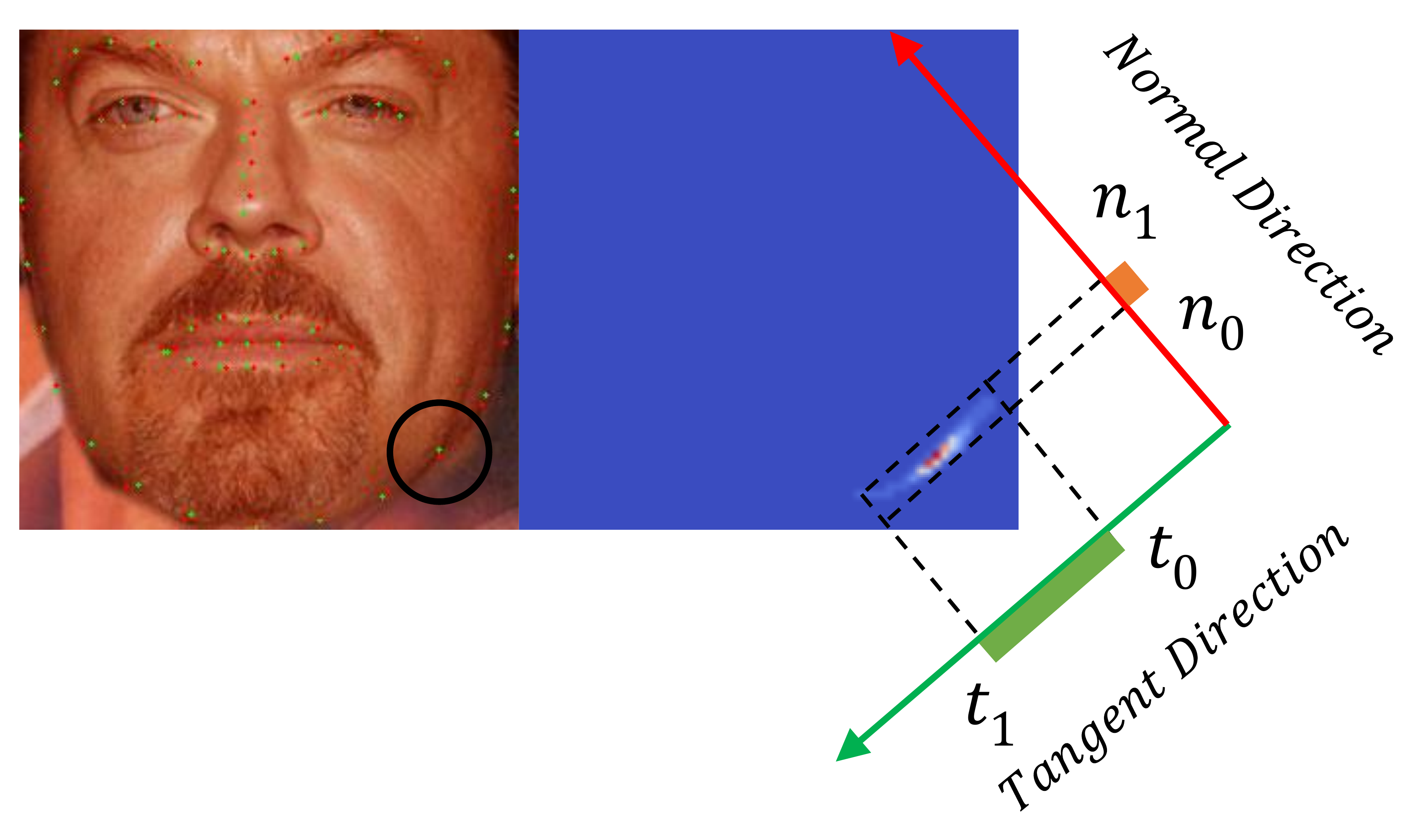}
\end{center}
\vspace{-5mm}
\caption{Error tolerance in different direction by applying AAM mask. The \textbf{\textcolor[rgb]{0.929,0.490,0.1921}{orange}} segment indicates the predicted coordinate range in normal direction, and \textbf{\textcolor[rgb]{0.4392,0.6784,0.2784}{green}} segment indicates the predicted coordinate range in tangent direction.}
\label{figure:AAM_mask}
\vspace{-5mm}
\end{figure}

As described in the manuscript, the anisotropic attention module outputs an anisotropic mask per landmarks.
By design, the anisotropic mask has a strong response in tangent direction and a weak response in normal direction.
Consequently, each predicted landmark has a large tolerance for tangent error, but small tolerance for normal error.
This can be confirmed in the visualized example in Figure~\ref{figure:AAM_mask}, where the AAM mask has broad distribution along tangent direction (ranging between $t_0$ to $t_1$) while the distribution along normal direction is limited (ranging between $n_0$ to $n_1$). 
In other words, the guideline imposes strong constraints along the normal direction of each landmark.

\begin{table*}[htbp]
\begin{center}
\begin{tabular}{|l|c|c|c|c|c|c|}
\hline
Components & Edges & Methods & Overall NME & Normal NME & Tangent NME & Bias Rate \\
\hline

\multirow{2}*{-} & \multirow{2}*{Overall} & Baseline & 3.38 & 1.91 & 2.55 & 33.51\% \\
 & & ADNet & 2.93 & 1.54 & 2.28 & 48.05\% \\
\hline

\multirow{2}*{Contour} & \multirow{2}*{Face Contour} & Baseline & 5.85 & 2.97 & 4.73 & 59.20\% \\
 & & ADNet & 5.45 & 2.58 & 4.57 & 77.13\% \\
\hline

\multirow{4}*{Eyebrow} & \multirow{2}*{Right Eyebrow} & Baseline & 3.62 & 2.10 & 2.75 & 30.51\% \\
 & & ADNet & 3.31 & 1.86 & 2.56 & 37.35\% \\
\cline{2-7}
 & \multirow{2}*{Left Eyebrow} & Baseline & 3.44 & 1.99 & 2.62 & 31.62\% \\
 & & ADNet & 3.15 & 1.75 & 2.45 & 40.24\% \\
\hline

\multirow{2}*{Nose} & \multirow{2}*{Nose Middle Line} & Baseline & 2.13 & 1.78 & 1.59 & 35.13\% \\
 & & ADNet & 1.97 & 1.01 & 1.53 & 51.03\% \\
\cline{2-7}
 & Nose Bottom Line & Baseline & 2.31 & 1.43 & 1.66 & 15.59\% \\
 & & ADNet & 2.11 & 1.26 & 1.56 & 23.27\% \\
\hline

\multirow{8}*{Eye} & \multirow{2}*{Right Eye Superior Margin} & Baseline & 1.88 & 1.23 & 1.25 & 1.83\% \\
 & & ADNet & 1.48 & 0.94 & 1.01 & 7.85\% \\
\cline{2-7}
 & \multirow{2}*{Right Eye Inferior Margin} & Baseline & 1.81 & 1.19 & 1.22 & 2.52\% \\
 & & ADNet & 1.42 & 0.89 & 0.98 & 10.11\% \\
\cline{2-7}
 & \multirow{2}*{Left Eye Superior Margin} & Baseline & 1.83 & 1.20 & 1.22 & 1.65\% \\
 & & ADNet & 1.43 & 0.92 & 0.96 & 3.96\% \\
\cline{2-7}
 & \multirow{2}*{Left Eye Inferior Margin} & Baseline & 1.80 & 1.17 & 1.20 & 2.56\% \\
 & & ADNet & 1.39 & 0.87 & 0.94 & 8.00\% \\
\hline

\multirow{8}*{Mouth} & \multirow{2}*{Outer Lip Superior Margin} & Baseline & 2.35 & 1.48 & 1.64 & 10.80\% \\
 & & ADNet & 2.01 & 1.18 & 1.47 & 24.25\% \\
\cline{2-7}
 & \multirow{2}*{Outer Lip Inferior Margin} & Baseline & 2.81 & 1.69 & 2.06 & 21.89\% \\
 & & ADNet & 2.62 & 1.52 & 1.98 & 30.26\% \\
\cline{2-7}
 & \multirow{2}*{Inner Lip Superior Margin} & Baseline & 2.15 & 1.32 & 1.49 & 12.61\% \\
 & & ADNet & 1.79 & 0.97 & 1.33 & 37.37\% \\
\cline{2-7}
 & \multirow{2}*{Inner Lip Inferior Margin} & Baseline & 2.48 & 1.53 & 1.79 & 16.99\% \\
 & & ADNet & 2.13 & 1.24 & 1.64 & 32.25\% \\
\hline

\end{tabular}
\end{center}
\caption{Evaluation of individual edges on 300W.}
\label{table:Edges_test_300W}
\end{table*}

\end{document}